\documentclass[10pt,twocolumn,letterpaper]{article}
\usepackage[pagenumbers]{cvpr}

\usepackage{amsmath,amsfonts,bm}

\def\eqref#1{equation~\ref{#1}}

\def\1{\bm{1}}

\DeclareMathAlphabet{\mathsfit}{\encodingdefault}{\sfdefault}{m}{sl}
\SetMathAlphabet{\mathsfit}{bold}{\encodingdefault}{\sfdefault}{bx}{n}

\usepackage{array}
\usepackage{makecell}
\usepackage{multirow}
\usepackage{tabularx}
\usepackage{colortbl}
\usepackage{nicefrac}
\usepackage{microtype}
\usepackage{placeins}
\newcommand{\methodname}{\texorpdfstring{P\textsuperscript{2}-TTA}{P2-TTA}}
\newsavebox{\cvprtablebox}
\newcommand{\fitwidth}[2]{%
  \begingroup
  \sbox{\cvprtablebox}{#2}%
  \ifdim\wd\cvprtablebox>#1\relax
    \resizebox{#1}{!}{\usebox{\cvprtablebox}}%
  \else
    \usebox{\cvprtablebox}%
  \fi
  \endgroup
}
\DeclareUnicodeCharacter{2193}{\ensuremath{\downarrow}}

\definecolor{cvprblue}{rgb}{0.21,0.49,0.74}
\usepackage[pagebackref,breaklinks,colorlinks,allcolors=cvprblue]{hyperref}
\hypersetup{
  pdftitle={Coordinating Perception and Planning for Test-Time Adaptation in End-to-End Driving under Distribution Shifts},
  pdfauthor={Shiyu Fang, Yiming Cui, Qihang Huang, Zhan Chen, Collister Chua, Wenhui Huang, Peng Hang, Chen Lv, Jian Sun}
}
\def\confName{CVPR}
\def\confYear{2027}
\newif\ifwidecomparisons
\widecomparisonsfalse
\title{Coordinating Perception and Planning for Test-Time Adaptation\\
in End-to-End Driving under Distribution Shifts}
\author{
  {\normalsize Shiyu Fang\textsuperscript{1}\quad Yiming Cui\textsuperscript{1}\quad Qihang Huang\textsuperscript{2}\quad Zhan Chen\textsuperscript{2}\quad Collister Chua\textsuperscript{2}}\\[3pt]
  {\normalsize Wenhui Huang\textsuperscript{2}\quad Peng Hang\textsuperscript{1}\quad Chen Lv\textsuperscript{2}\quad Jian Sun\textsuperscript{1}}\\[5pt]
  {\normalsize\textsuperscript{1}Tongji University\qquad\textsuperscript{2}Nanyang Technological University}
}

\begin{document}
\maketitle
\begin{abstract}
End-to-end autonomous driving has achieved significant milestones, yet its reliance on the i.i.d. assumption renders it vulnerable to real-world distribution shifts. While Test-Time Adaptation (TTA) offers a promising paradigm for on-the-fly enhancement, its application to driving is hindered by two barriers: (i) \textit{multi-task adaptation dependency}, where updates to upstream perception also affect downstream planning, while perception confidence alone does not directly assess these downstream effects; and (ii) \textit{continuous adaptation instability}, where recursive online updates over long horizons cause compounding parameter drift. To address these challenges, we propose \methodname{}, a plug-and-play TTA framework for the continual adaptation of end-to-end driving models. \methodname{} coordinates adaptation through a joint objective combining perception entropy minimization with a retrospective temporal planning consistency objective derived from observed ego motion. {To stabilize online updates, we use an EMA teacher as a moving parameter reference and reject candidate updates that increase historical planning loss.} {Extensive experiments across multiple models demonstrate that, by updating {an average of 0.11\textperthousand\ of parameters with an average adaptation time of 19.9\,ms on different backbones}, {\methodname{}} achieves average recovery rates of $36.4\%$, $30.0\%$, and {$9.3\%$} for L2, collision rate, and PDMS, respectively, under severe distribution shifts. {Under a shared perception--planning objective,} {\methodname{}} achieves the highest L2 and collision recovery among the compared TTA baselines on UniAD.}
\end{abstract}
\section{Introduction}
\begin{figure*}[t]
  \centering
  \includegraphics[width=\linewidth]{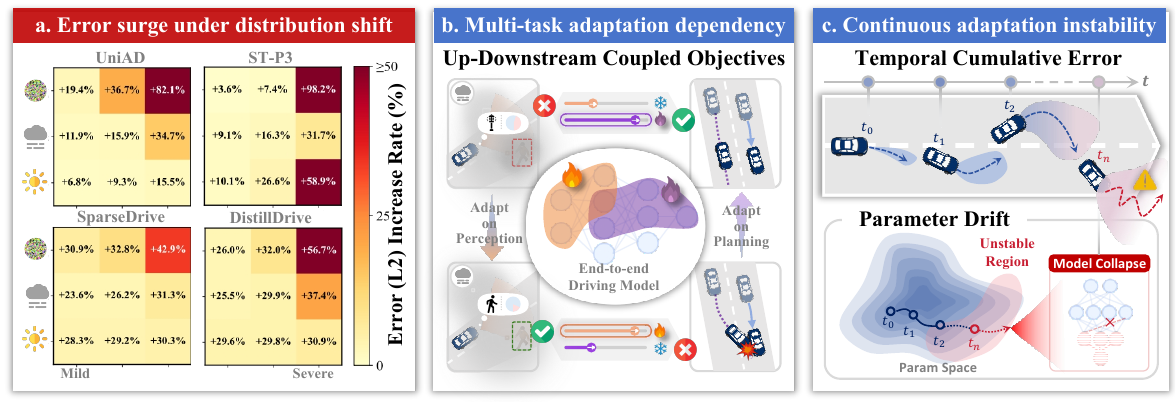}
  \caption{\textbf{Motivation and core challenges.} Distribution shifts cause severe performance degradation in {end-to-end autonomous driving models}, \textcolor{black}{motivating adaptation that accounts for perception--planning dependencies and temporal stability.}}
  \label{fig:motivation}
\end{figure*}

End-to-end autonomous driving has emerged as a prevalent paradigm, demonstrating remarkable proficiency and generalization potential across diverse driving scenarios \citep{chen2024end, fang2025corevla}. However, the performance of such models relies heavily on the assumption that test and training data are independently and identically distributed (i.i.d.), which rarely holds in practice \citep{do2024reducing, shoeb2025out}. As shown in {Figure~\ref{fig:motivation}(a)}, these models degrade sharply under various real-world distribution shifts; for instance, {dense fog, brightness fluctuations, or Gaussian noise can cause error rates to surge under severe corruption}. \textcolor{black}{Existing approaches address distribution shifts through domain adaptation (DA) \citep{ben2006analysis, ning2025multilevel} or data augmentation \citep{zhong2020random, wang2021augmax}, but typically require access to target-domain data in advance, additional training, or a predefined corruption model.} \textcolor{black}{These requirements limit their applicability when test conditions change unexpectedly \citep{malinin2021shifts, diehl2025lord}.} This motivates equipping end-to-end autonomous driving models with the ability to adapt online at test time, requiring neither {labeled data} nor offline retraining.

Test-time adaptation (TTA) offers a compelling alternative by updating {models} on the fly using only unlabeled test samples \citep{wang2020tent, liang2025comprehensive}. Early fully TTA (FTTA) methods adapt the model via lightweight self-supervised objectives \citep{shin2024tta, xiao2024beyond, karmanov2024efficient}. Continual TTA (ConTTA) methods further extend this by continuing to update model parameters across successive batches of test samples, introducing mechanisms to prevent forgetting \citep{park2025hybrid, hoang2024persistent, tan2025uncertainty}. \textcolor{black}{Many established TTA methods have been developed for visual recognition, particularly image classification, where adaptation is usually guided by task-specific prediction signals.} \textcolor{black}{Applying such methods to end-to-end driving additionally requires considering the influence of perceptual updates on planning and the temporal dependence of successive observations.}

Adapting end-to-end driving models at test time poses two fundamental challenges for existing TTA methods.
(i) \textit{Multi-task adaptation dependency.}
{Existing entropy-based TTA methods primarily target image classification, where prediction confidence guides adaptation \citep{wang2020tent, niu2022efficient, niu2023towards}. In end-to-end driving, upstream perception and downstream planning are coupled: degraded perceptual representations under distribution shifts can propagate to planning outputs, causing predicted ego motion to deviate from subsequently observed vehicle motion. }
(ii) \textit{Continuous adaptation instability.}
Unlike conventional image classification, where test samples are often treated as independent prediction instances, autonomous driving operates on temporally correlated observation streams. Consequently, successive online updates may accumulate errors over time, increasing the risk of parameter drift.

To address these challenges, we propose {Perception--Planning Test-Time Adaptation (\textbf{\methodname{}})}, a plug-and-play continual adaptation framework for end-to-end driving, enabling models to continuously update parameters during inference.
{To coordinate these coupled tasks, \methodname{} combines perception entropy with temporal planning consistency in a joint adaptation objective. The two signals provide complementary guidance from perception confidence and observed ego motion.}
{\color{black}{To mitigate unstable parameter changes, \methodname{} projects candidate updates around an EMA parameter reference and accepts them only when a retrospective planning consistency criterion improves. Together, these mechanisms constrain parameter changes during online adaptation.}}
Extensive experiments on {nuScenes and NAVSIM} under various corruptions demonstrate that {\methodname{}} generalizes across various end-to-end models of diverse architectures, achieving the highest L2 and collision recovery among the compared TTA baselines on UniAD. Moreover, ablation studies further validate the contribution of each component, {while efficiency analysis shows low additional adaptation overhead}. In summary, our contributions are as follows:
\begin{itemize}[leftmargin=*]
\item {We propose {\methodname{}}, a plug-and-play framework for end-to-end driving that enables diverse pretrained models to adapt at test time under severe distribution shifts.}
\item {We formulate a joint self-supervised objective that integrates upstream perception entropy with retrospective temporal planning consistency, guiding model updates toward more reliable end-to-end driving under distribution shifts.}
\item {We introduce a teacher-student architecture that combines EMA parameter anchoring with per-step update verification to constrain parameter changes during adaptation.}
\item {We demonstrate that {\methodname{}} updates {an average of 0.11\textperthousand\ of parameters with an average adaptation time of $19.9$\,ms}, achieving average recovery rates of $36.4\%$, $30.0\%$, and {$9.3\%$} for L2, collision rate, and PDMS, respectively, under distribution shift.}
\end{itemize}

\section{Related Work}\label{sec:2}
\textbf{Domain Adaptation.} DA addresses distribution shift by aligning features between the labeled source domain and the unlabeled target domain \citep{farahani2021brief, sohn2017unsupervised, you2019universal, kodirov2015unsupervised}, typically through moment matching \citep{peng2019moment}, adversarial domain discrimination \citep{ganin2016domain, tang2020discriminative, he2025unsupervised}, or optimal transport \citep{courty2016optimal}, achieving strong performance when target data is available at training time. However, DA methods require access to target domain data during training and often necessitate retraining—making them impractical when the test distribution is unknown in advance.

\textbf{Test-Time Training.}
Test-Time Training (TTT) adapts models using self-supervised objectives on unlabeled test inputs, without requiring target-domain data in advance \citep{sun2020test, dalal2025one}.
Traditional approaches employ auxiliary tasks, such as rotation prediction \citep{sun2020test} or noise-contrastive learning \citep{osowiechi2024nc}, which often require modifications to the training pipeline.
More recently, Centaur \citep{centaur2025} extends TTT to end-to-end driving by minimizing cluster entropy over direction-grouped candidate trajectories.
However, existing approaches primarily rely on task-specific adaptation signals without explicitly accounting for the coupling between upstream perception and downstream planning.

\textbf{Fully Test-Time Adaptation.} FTTA requires no modification to training, updating model parameters using unsupervised signals at test time \citep{wang2020tent, zhao2023delta, rafi2026learning, chu2025remember, niu2022efficient}. Entropy minimization over model predictions has emerged as the dominant paradigm, with subsequent work introducing various mechanisms to improve update reliability and prevent parameter drift \citep{tan2025uncertainty, zhang2022auxadapt, hoang2024persistent}. Beyond entropy minimization, contrastive learning combined with pseudo-labeling also provides adaptation signals \citep{chen2022contrastive}. Sustaining reliable adaptation across sequential inputs is studied in continual TTA.

\textbf{Continual Test-Time Adaptation.} ConTTA extends FTTA to settings where the model must update over a sequence of test samples, requiring adaptation without forgetting \citep{niu2023towards, wang2022continual, song2023ecotta}. To prevent error accumulation, recent methods maintain an exponential moving average (EMA) model to generate stable pseudo-targets and stochastically restore a subset of parameters toward their pretrained values to preserve source knowledge \citep{sojka2023ar, ye2025domain, tan2026exposing, chen2024cross}. While these methods address sequential distribution shifts, they do not necessarily provide a planning-related supervision signal tailored to end-to-end driving. In particular, preserving parameter stability and improving downstream trajectory quality are distinct objectives that should be evaluated separately.

\section{Problem Definition}\label{sec:3}

\begin{table}[t]
\caption{Comparison of adaptation paradigms. Source Acc. denotes access to source-domain data during adaptation.
Task and Frame denote cross-task and cross-frame adaptation. $x$ is an unlabeled test sample; $\mathcal{L}_{sup}$, $\mathcal{L}_{align}$, $\mathcal{L}_{aux}$, and $\mathcal{L}_{self}$ denote supervised, alignment, auxiliary, and self-supervised objectives. Shading marks {\methodname{}}.}
\label{tab:tta}
\centering
\small
\setlength{\tabcolsep}{2pt}
\resizebox{\linewidth}{!}{%
\begin{tabular}{@{}l|ccccc@{}}
\toprule
Method & Phase & Objective & Source Acc. & Task & Frame \\
\midrule
DA & Train & $\mathcal{L}_{sup}+\mathcal{L}_{align}$ & \checkmark & - & - \\
TTT & Test + Train & $\mathcal{L}_{sup}+\mathcal{L}_{aux}(x^t)$ & - & - & - \\
FTTA & Test & $\mathcal{L}_{self}(x^t)$ & - & - & - \\
ConTTA & Test & $\mathcal{L}_{self}(x^t)$ & \checkmark & - & - \\
\rowcolor{blue!5}{\methodname{}} (Ours) & Test & $\mathcal{L}_{self}(x^t)$ & - & \checkmark & \checkmark \\
\bottomrule
\end{tabular}}
\end{table}

Table~\ref{tab:tta} summarizes existing adaptation paradigms and highlights the two properties that existing TTA methods leave unaddressed. Current TTA methods share a common update rule: given an unlabeled test sample $x_s$ at step $s$, the affine normalization parameters in BatchNorm $\theta^{BN}$ are updated by minimizing a self-supervised objective $\mathcal{L}_{self}$,

\noindent
\begin{equation}
\theta_{s}^{BN} = \theta_{s-1}^{BN} - \eta 
\nabla_{\theta} \mathcal{L}_{self}({x_s};\, \theta_{s-1}^{BN}),
\label{eq:tta_bn_update}
\end{equation}
where $\eta$ is the learning rate. Unlike a single-task predictor, an end-to-end driving model maps its input through perception representations $\mathcal{Z}_{perc}$ to a downstream planning output $\hat{\mathcal{Y}}_{plan}$ via a mapping $f_{prog}$. Since autonomous driving processes inputs as a continuous sequence $x_t$ indexed by time $t$, the adaptation naturally targets LayerNorm parameters $\theta^{LN}$, and the upstream-to-downstream dependency takes the form:
\begin{equation}
\hat{\mathcal{Y}}_{plan} = f_{prog}\big(x_t \mid
\mathcal{Z}_{perc}(x_t;\, \theta^{LN}),\, \theta^{LN}\big).
\label{eq:perception_to_planning}
\end{equation}
Under distribution shift, corruption in $\mathcal{Z}_{perc}$ propagates through $f_{prog}$, causing errors to cascade from perception to planning; conversely, shared parameters couple gradient signals back across tasks.

{\color{black}Continual adaptation retains parameter changes across successive test inputs, but reliable driving adaptation also needs a planning-related signal derived from past and current observations. We denote the selected LayerNorm affine parameters collectively by $\theta_t \equiv \theta_t^{LN}$ and the EMA parameter reference by $\theta_t^*$. The proposed adaptation step consists of a candidate optimizer update followed by stabilization:}
\begin{equation}
\color{black}
\begin{aligned}
g_t &= \left.\nabla_{\theta}\mathcal{L}_{\mathrm{joint},t}(\theta)\right|_{\theta=\theta_{t-1}}, \\
\theta'_t &= \operatorname{OptStep}(\theta_{t-1},g_t;\eta), \\
\theta_t &= \mathcal{S}_t(\theta'_t;\theta_{t-1},\theta^*_{t-1}).
\end{aligned}
\label{eq:unified_adaptation}
\color{black}
\end{equation}
\textcolor{black}{Here $g_t$ is the gradient of the joint objective at the pre-update parameters; $\operatorname{OptStep}$ denotes one optimizer proposal (Adam in our experiments, with its internal optimizer state); and $\mathcal{S}_t$ denotes the data-dependent stabilization procedure consisting of projection and update acceptance, specified in Sec.~\ref{sec:4}.} \textcolor{black}{The time-indexed joint loss is defined below.} \textcolor{black}{The stabilization procedure uses only observations available by time $t$.}

\section{Methodology}\label{sec:4}

To address the cross-task and cross-frame requirements in Eq.~\ref{eq:unified_adaptation}, {\methodname{}} introduces two complementary components. Taking into account the coupling relationship between upstream perception and downstream planning, {\methodname{}} constructs a joint self-supervised objective $\mathcal{L}_{joint}$ to enable unified cross-task coordination throughout the adaptation process.
{A teacher-student framework additionally constrains parameter updates and verifies them using temporal planning consistency.}
{\methodname{} updates only the affine parameters $(\gamma,\beta)$ of selected LayerNorm modules. Given the architectural differences across driving models, the LayerNorm modules selected for adaptation vary accordingly. Model-specific LayerNorm configurations are detailed in Appendix~\ref{app:f}.}

\subsection{{Joint Objective for Perception--Planning Coordination}}
End-to-end autonomous driving models are inherently multi-task systems with highly interdependent components \citep{grigorescu2020survey, chitta2022transfuser, huang2023gameformer}. {During supervised training, task-specific labels provide reference signals for optimizing perception and planning. At test time, however, such labels are unavailable, motivating a self-supervised adaptation objective that guides online parameter updates using unlabeled observations. While minimizing perception entropy encourages confident predictions, it does not directly assess the effect of an update on downstream planning. We therefore combine this upstream signal with a temporal planning consistency loss based on observed ego motion, enabling coordinated adaptation of perception and planning:}
\begin{equation}
\color{black}
\mathcal{L}_{\mathrm{joint},t}(\theta)
= \mathcal{L}_{\mathrm{mm},t}(\theta)
+ \lambda\mathcal{L}_{\mathrm{temp},t}(\theta),
\label{eq:joint_objective}
\color{black}
\end{equation}
\textcolor{black}{Here $\mathcal{L}_{\mathrm{mm},t}(\theta)$ measures perception-side uncertainty and $\mathcal{L}_{\mathrm{temp},t}(\theta)$ measures retrospective planning consistency using ego motion observed at time $t$; $\lambda$ controls their relative weight.} \textcolor{black}{When the time step and parameters are clear, we abbreviate these terms as $\mathcal{L}_{mm}$, $\mathcal{L}_{temp}$, and $\mathcal{L}_{joint}$.} \textcolor{black}{Both objectives are evaluated using the same trainable LayerNorm parameter vector $\theta$.}

\textbf{Upstream Perception: Multi-modal Entropy Minimization.} 
Under severe environmental perturbations—such as dense fog or sensor noise—the model’s perceptual discriminability degrades, causing the output probability distributions to undergo a flattening effect that manifests as heightened predictive uncertainty \citep{hendrycks2019benchmarking, ovadia2019can}. This flattening indicates that the latent embeddings have drifted into low-density regions of the source-domain manifold, making it difficult for the model to produce decisive categorical or attribute assignments.

To quantify the uncertainty arising from the distortion of $\mathcal{Z}_{perc}$, we formalize the output process of the perception heads. At each time step $t$, the system generates $N$ parallel perceptual responses (e.g., object detection, semantic occupancy) based on the latent features, where the normalized confidence $q_{t,i}$ for the $i$-th response is mapped via: 
\begin{equation}
\color{black}
q_{t,i}(\theta) = \sigma\!\left(\Phi^{(i)}\!\left(\mathcal{Z}_{perc}(x_t,\mathcal{H}_{t-1};\theta)\right)\right)
\color{black}
\end{equation}
\textcolor{black}{Here $\sigma(\cdot)$ is the sigmoid function, $\Phi^{(i)}$ denotes the selected perceptual output mapping, and $\mathcal{H}_{t-1}$ is the historical context available at time $t$.} \textcolor{black}{The formulation below applies to the selected Bernoulli-type probability outputs.} \textcolor{black}{For each such output, we use binary entropy as an uncertainty measure:}
\begin{equation}
\color{black}
\mathcal{L}_{\mathrm{mm},t}(\theta) = \frac{1}{N} \sum_{i=1}^{N} \mathbb{E}_{ent}(q_{t,i}(\theta))
\color{black}
\end{equation}
{\color{black}where $\mathbb{E}_{\mathrm{ent}}(q) = -q\log q - (1-q)\log(1-q)$. Minimizing $\mathcal{L}_{\mathrm{mm},t}$ encourages confident perceptual outputs, although confidence alone does not guarantee semantic correctness under distribution shift.}

\textbf{Downstream Planning: Temporal Consistency Optimization.} 
Crucially, solely minimizing perceptual entropy may lead to decisive but erroneous interpretations, as the model might confidently converge toward incorrect scene semantics. Moreover, under distribution shifts, degraded perceptual representations can propagate to planning outputs, causing the ego position predicted from a previous observation to deviate from the position subsequently observed. To provide a planning-related feedback signal without requiring future trajectory labels, we exploit the temporal correspondence between consecutive observations: the ego position predicted from the previous-frame input for the current time should agree with the ego position observed at the current time.

\textcolor{black}{At time $t$, let $\mathcal{I}_t^{\mathrm{hist}}$ denote the available historical planning input and its context. We re-evaluate the corresponding ego-position prediction under parameters $\theta$:}
\begin{equation}
\color{black}
\hat p_t^{\mathrm{hist}}(\theta)=f_{plan}(\mathcal{I}_t^{\mathrm{hist}};\theta).
\label{eq:retro_prediction}
\color{black}
\end{equation}
{\color{black}The prediction and the corresponding observed ego position $p_t^{\mathrm{obs}}$ are expressed in a common coordinate frame. Their consistency is measured by}
\begin{equation}
\color{black}
\mathcal{L}_{\mathrm{temp},t}(\theta)
=\big\|\hat p_t^{\mathrm{hist}}(\theta)-p_t^{\mathrm{obs}}\big\|_2^2.
\label{eq:l_temp}
\color{black}
\end{equation}
\textcolor{black}{The historical forward pass is re-evaluated as $\theta$ changes, while its inputs and observation reference remain fixed across parameter candidates. Together with perception entropy, this observed-motion constraint provides complementary guidance for coordinating perception and planning during adaptation.}

\subsection{Teacher-Student Architecture for Cross-frame Stability} 

Autonomous driving is inherently a continuous temporal process where environmental shifts are not isolated events but persistent perturbations. Despite the synergetic guidance imposed by our joint objective $\mathcal{L}_{joint}$, adaptation over continuous test streams faces the risk of error accumulation. Since each update is conditioned on previously adapted states, minor gradient biases at a single frame can gradually compound, leading to severe parameter drift or even catastrophic model collapse.

\textcolor{black}{To constrain online updates, we use a teacher--student stabilization procedure $\mathcal{S}_t$ composed of EMA-centered parameter projection and retrospective consistency-based update acceptance \citep{gong2022note, dobler2023robust, lee2023tta}.} \textcolor{black}{The EMA parameters $\theta^*_t$ provide a moving reference; the student moves through the states $\theta_{t-1} \to \theta'_t \to \hat\theta_t \to \theta_t$, representing the pre-update, optimizer-proposed, projected, and accepted-or-reverted parameter values.} \textcolor{black}{At the start of an independently adapted sequence, the EMA reference is initialized as $\theta^*_0=\theta_0$.}

\textbf{Student: Candidate Parameter Update.}
\textcolor{black}{At time $t$, the student computes the gradient $g_t$ of the adaptation objective, as defined in Eq.~\ref{eq:unified_adaptation}, and proposes an optimizer update:}
\begin{equation}
\color{black}
\theta'_t = \operatorname{OptStep}(\theta_{t-1},g_t;\eta).
\label{eq:candidate_update}
\color{black}
\end{equation}
\textcolor{black}{The operator $\operatorname{OptStep}$ represents one optimizer update, implemented with Adam in the experiments and including its internal optimizer state.} \textcolor{black}{The proposed parameters are not yet finalized: projection and the acceptance test follow before they can be used as the next student state.}

\textbf{Teacher: Two-scale Stability Guidance.}
{During continuous online adaptation, successive parameter updates may accumulate errors over time, increasing the risk of parameter drift. We therefore use an EMA reference to constrain candidate parameter updates and a consistency check against the previous-step planning prediction to determine whether each update should be retained.}

\textcolor{black}{The EMA reference tracks the history of finalized student parameter values without participating in backpropagation.} \textcolor{black}{It is updated only after the acceptance decision for time $t$:}
\begin{equation}
\theta^{*}_{t} = \alpha \theta^{*}_{t-1} + (1 - \alpha) \theta_{t},
\label{eq:ema_update}
\end{equation}
\textcolor{black}{where $\alpha$ controls the smoothing rate.} \textcolor{black}{The resulting teacher is a moving reference, rather than a fixed constraint around the pretrained initialization.}

{\color{black}Before accepting an update, the proposed student parameters are projected into an $L_\infty$ ball of radius $r$ centered on the EMA reference from the previous step:}
\begin{equation}
\color{black}
\hat{\theta}_{t} = \Pi_{\mathcal{B}_{\infty}(\theta^{*}_{t-1}, r)}(\theta'_{t}).
\label{eq:projection}
\color{black}
\end{equation}
\textcolor{black}{Here $\Pi$ performs elementwise clipping of the selected trainable parameters around $\theta^*_{t-1}$.} \textcolor{black}{The constraint bounds the deviation of each candidate from the moving teacher, but does not impose a global bound relative to the original pretrained weights.}

{\color{black}After projection, a consistency-based acceptance check compares the same retrospective loss at the projected candidate and at the pre-update student parameters. Both evaluations use $\mathcal{I}_t^{\mathrm{hist}}$ and $p_t^{\mathrm{obs}}$ from the same adaptation step; thus the comparison isolates the effect of the candidate parameter values rather than a change in evaluation inputs:}
\begin{equation}
\color{black}
\theta_{t}=
\begin{cases}
\hat{\theta}_t,
& \mathcal{L}_{\mathrm{temp},t}(\hat{\theta}_t)
 < \mathcal{L}_{\mathrm{temp},t}(\theta_{t-1}),\\
\theta_{t-1}, & \text{otherwise}.
\end{cases}
\label{eq:acceptance}
\color{black}
\end{equation}
\textcolor{black}{The decision is based on the retrospective planning-consistency proxy, not on ground-truth future trajectories or direct evaluation of collision safety.} \textcolor{black}{At each step, the student first proposes an optimizer update, then projects it using the previous EMA reference, applies the acceptance check, and finally updates the EMA teacher from the finalized $\theta_t$ according to Eq.~\ref{eq:ema_update}.} \textcolor{black}{Rejected updates restore the pre-update parameters and Adam state. The next planning prediction uses the finalized student parameters.}

\section{Experiments}
\subsection{Experimental Setup}
{\textcolor{black}{We evaluate \methodname{} on nuScenes and NAVSIM using officially released pretrained weights under two distinct adaptation protocols.} \textcolor{black}{On nuScenes, parameters are updated sequentially across recorded frames within a scene and reset at scene boundaries; the observed ego state is available only up to the current frame, and the evaluated planner does not control the recorded motion.} \textcolor{black}{We report L2 trajectory error and collision rate averaged over the 1s, 2s, and 3s prediction horizons.} \textcolor{black}{On NAVSIM, each independently evaluated sample uses three observations for a single adaptation step before its predicted plan is assessed by a non-reactive rollout, and parameters are reset between samples.} \textcolor{black}{We report PDMS for NAVSIM.} \textcolor{black}{In both protocols, $p_t^{\mathrm{obs}}$ is drawn from recorded ego-state observations available no later than time $t$, without using future trajectory labels for adaptation.} Recovery rate (RR) measures the fraction of corruption-induced degradation recovered relative to the unadapted clean and noisy references; mean RR averages per-model rates with positive clean-to-noisy degradation. Unless otherwise specified, adaptation uses Adam with learning rate $\eta=0.01$, EMA momentum $\alpha=0.8$, projection radius $r=0.005$, and loss weight $\lambda=1$. The default hyperparameters are shared across corruption types and remain fixed at test time. All experiments are conducted on NVIDIA A800 GPUs. All reported results are averaged over three runs with different random seeds.}

\subsection{Main Results}

\begin{table}[!b]
\caption{\textbf{nuScenes performance under severity-5 Gaussian noise.}
Results at severity~5, averaged over 1s, 2s, and 3s {(AVG)}. SparseDrive reports endpoint scores on frames with a complete 3s future, with matched Clean/Noisy references. {Shading marks TTA results; red values report recovery rates computed before display rounding.}}
\label{tab:main}
\centering
\small
\setlength{\tabcolsep}{2pt}
\resizebox{\linewidth}{!}{%
\begin{tabular}{@{}l|ccc|ccc@{}}
\toprule
\multirow{2}{*}{Model} & \multicolumn{3}{c|}{L2 (m)$\downarrow$} & \multicolumn{3}{c}{Col. (\%)$\downarrow$} \\
& Clean & Noisy & TTA & Clean & Noisy & TTA \\
\midrule
ST-P3 \citep{hu2022st} & 3.28 & 5.29 & \cellcolor{blue!5}\textbf{4.60}{\scriptsize \textcolor{red}{(+34.3\%)}} & 1.61 & 5.13 & \cellcolor{blue!5}\textbf{4.03}{\scriptsize \textcolor{red}{(+31.3\%)}} \\
UniAD \citep{hu2023planning} & 0.94 & 1.38 & \cellcolor{blue!5}\textbf{1.25}{\scriptsize \textcolor{red}{(+29.5\%)}} & 0.29 & 0.95 & \cellcolor{blue!5}\textbf{0.72}{\scriptsize \textcolor{red}{(+34.8\%)}} \\
DistillDrive \citep{yu2025distilldrive} & 0.58 & 0.80 & \cellcolor{blue!5}\textbf{0.76}{\scriptsize \textcolor{red}{(+18.2\%)}} & 0.07 & 0.31 & \cellcolor{blue!5}\textbf{0.28}{\scriptsize \textcolor{red}{(+12.5\%)}} \\
SparseDrive \citep{sun2025sparsedrive} & 1.13 & 1.50 & \cellcolor{blue!5}\textbf{1.40}{\scriptsize \textcolor{red}{(+26.5\%)}} & 0.30 & 1.30 & \cellcolor{blue!5}\textbf{0.85}{\scriptsize \textcolor{red}{(+45.5\%)}} \\
DiffusionDrive \citep{liao2025diffusiondrive} & 0.57 & 0.77 & \cellcolor{blue!5}\textbf{0.74}{\scriptsize \textcolor{red}{(+15.0\%)}} & 0.08 & 0.29 & \cellcolor{blue!5}\textbf{0.23}{\scriptsize \textcolor{red}{(+28.6\%)}} \\
CoIRL-AD \citep{coirl} & 0.63 & 0.65 & \cellcolor{blue!5}\textbf{0.63}{\scriptsize \textcolor{red}{(+94.8\%)}} & 0.18 & 0.25 & \cellcolor{blue!5}\textbf{0.23}{\scriptsize \textcolor{red}{(+27.3\%)}} \\
\bottomrule
\end{tabular}}
\end{table}

{Under severity-5 Gaussian noise, Table~\ref{tab:main} shows mean recovery rates of $36.4\%$ for L2 and $30.0\%$ for collision rate across six end-to-end driving architectures. These gains span different modeling paradigms: the unified multi-task framework UniAD~\citep{hu2023planning} recovers $29.5\%$ and $34.8\%$ of the L2 and collision-rate degradation, respectively; and the diffusion-based planner DiffusionDrive~\citep{liao2025diffusiondrive} achieves $15.0\%$ and $28.6\%$. These examples support the applicability of our adaptation framework across multiple driving models under the evaluated distribution shift.}

\begin{table*}[!t]

\caption{{\textcolor{black}{\textbf{Planning performance under visual and physical perturbations.}}
Fog, snow, and brightness corruptions are evaluated at severity level 5. Physical combines GPS noise (1\,m), yaw noise ($2^{\circ}$), and motion blur.}}
\label{tab:corruptions}
\centering
\small
\setlength{\tabcolsep}{1pt}
\renewcommand{\arraystretch}{1.1}
\resizebox{\linewidth}{!}{%
\begin{tabular}{@{}l|cccc|cccc|cccc|cccc@{}}
\toprule
\multirow{3}{*}{Model} & \multicolumn{4}{c|}{Fog} & \multicolumn{4}{c|}{Snow} & \multicolumn{4}{c|}{Brightness} & \multicolumn{4}{c}{Physical} \\
\cmidrule(lr){2-5}\cmidrule(lr){6-9}\cmidrule(lr){10-13}\cmidrule(lr){14-17}
& \multicolumn{2}{c}{L2 (m)$\downarrow$} & \multicolumn{2}{c|}{Col. (\%)$\downarrow$} & \multicolumn{2}{c}{L2 (m)$\downarrow$} & \multicolumn{2}{c|}{Col. (\%)$\downarrow$} & \multicolumn{2}{c}{L2 (m)$\downarrow$} & \multicolumn{2}{c|}{Col. (\%)$\downarrow$} & \multicolumn{2}{c}{L2 (m)$\downarrow$} & \multicolumn{2}{c}{Col. (\%)$\downarrow$} \\
& Noisy & TTA & Noisy & TTA & Noisy & TTA & Noisy & TTA & Noisy & TTA & Noisy & TTA & Noisy & TTA & Noisy & TTA \\
\midrule
ST-P3 \citep{hu2022st} & 5.77 & \textbf{3.78}{\scriptsize (\textcolor{red}{+79.9\%})} & 5.38 & \textbf{3.15}{\scriptsize (\textcolor{red}{+59.2\%})} & 7.69 & \textbf{6.54}{\scriptsize (\textcolor{red}{+26.1\%})} & 6.58 & \textbf{3.43}{\scriptsize (\textcolor{red}{+63.4\%})} & 4.06 & \textbf{3.76}{\scriptsize (\textcolor{red}{+38.5\%})} & 2.42 & \textbf{2.11}{\scriptsize (\textcolor{red}{+38.3\%})} & 3.94 & \textbf{3.82}{\scriptsize (\textcolor{red}{+18.6\%})} & 3.08 & \textbf{2.35}{\scriptsize (\textcolor{red}{+49.3\%})} \\
UniAD \citep{hu2023planning} & 1.03 & \textbf{1.00}{\scriptsize (\textcolor{red}{+33.3\%})} & 0.32 & \textbf{0.31}{\scriptsize (\textcolor{red}{+33.3\%})} & 1.18 & \textbf{1.13}{\scriptsize (\textcolor{red}{+20.8\%})} & 0.97 & \textbf{0.84}{\scriptsize (\textcolor{red}{+19.1\%})} & 0.98 & \textbf{0.97}{\scriptsize (\textcolor{red}{+34.0\%})} & 0.33 & \textbf{0.32}{\scriptsize (\textcolor{red}{+18.2\%})} & 1.68 & \textbf{1.25}{\scriptsize (\textcolor{red}{+57.5\%})} & 1.30 & \textbf{0.32}{\scriptsize (\textcolor{red}{+97.4\%})} \\
DistillDrive \citep{yu2025distilldrive} & 0.63 & \textbf{0.61}{\scriptsize (\textcolor{red}{+40.0\%})} & 0.12 & \textbf{0.11}{\scriptsize (\textcolor{red}{+20.0\%})} & 0.97 & \textbf{0.87}{\scriptsize (\textcolor{red}{+25.6\%})} & 0.44 & \textbf{0.42}{\scriptsize (\textcolor{red}{+5.4\%})} & 0.60 & \textbf{0.60}{\scriptsize (\textcolor{red}{+21.6\%})} & 0.12 & \textbf{0.09}{\scriptsize (\textcolor{red}{+62.5\%})} & 1.14 & \textbf{1.13}{\scriptsize (\textcolor{red}{+2.4\%})} & 0.77 & \textbf{0.70}{\scriptsize (\textcolor{red}{+9.1\%})} \\
SparseDrive \citep{sun2025sparsedrive} & 0.63 & \textbf{0.62}{\scriptsize (\textcolor{red}{+21.1\%})} & 0.12 & \textbf{0.11}{\scriptsize (\textcolor{red}{+25.0\%})} & 0.85 & \textbf{0.80}{\scriptsize (\textcolor{red}{+22.7\%})} & 0.49 & \textbf{0.48}{\scriptsize (\textcolor{red}{+2.4\%})} & 0.82 & \textbf{0.75}{\scriptsize (\textcolor{red}{+36.6\%})} & 0.20 & \textbf{0.18}{\scriptsize (\textcolor{red}{+16.5\%})} & 1.26 & \textbf{1.21}{\scriptsize (\textcolor{red}{+8.9\%})} & 0.76 & \textbf{0.61}{\scriptsize (\textcolor{red}{+21.2\%})} \\
DiffusionDrive \citep{liao2025diffusiondrive} & 0.62 & \textbf{0.58}{\scriptsize (\textcolor{red}{+80.0\%})} & 0.12 & \textbf{0.12}{\scriptsize (\textcolor{red}{0.0\%})} & 1.03 & \textbf{0.94}{\scriptsize (\textcolor{red}{+19.6\%})} & 0.40 & \textbf{0.37}{\scriptsize (\textcolor{red}{+9.4\%})} & 0.75 & \textbf{0.72}{\scriptsize (\textcolor{red}{+16.3\%})} & 0.20 & \textbf{0.13}{\scriptsize (\textcolor{red}{+57.9\%})} & 1.00 & \textbf{0.99}{\scriptsize (\textcolor{red}{+2.3\%})} & 0.26 & \textbf{0.25}{\scriptsize (\textcolor{red}{+5.5\%})} \\
CoIRL-AD \citep{coirl} & 0.64 & \textbf{0.63}{\scriptsize (\textcolor{red}{+91.1\%})} & 0.36 & \textbf{0.35}{\scriptsize (\textcolor{red}{+2.4\%})} & 0.64 & \textbf{0.64}{\scriptsize (\textcolor{red}{+41.0\%})} & 0.53 & \textbf{0.46}{\scriptsize (\textcolor{red}{+19.6\%})} & 0.66 & \textbf{0.66}{\scriptsize (\textcolor{red}{+1.3\%})} & 0.26 & \textbf{0.26}{\scriptsize (\textcolor{red}{0.0\%})} & 0.63 & \textbf{0.63}{\scriptsize (\textcolor{red}{+42.5\%})} & 0.19 & \textbf{0.19}{\scriptsize (\textcolor{red}{0.0\%})} \\
\bottomrule
\end{tabular}}
\end{table*}

{To further evaluate the robustness of \methodname{} against a broader range of disturbances that may arise in real-world driving, we extend our evaluation beyond Gaussian noise to fog, snow, and brightness changes. Fog and snow primarily reduce scene visibility, whereas brightness changes alter illumination conditions. These visual corruptions are generated using the imagecorruptions library. Beyond these visual corruptions, localization errors can also affect the reliability of driving-model inputs. We therefore introduce a Physical setting that combines GPS position noise, yaw noise, and visual motion blur, simulating compound disturbances affecting both ego-state observations and camera inputs. Table~\ref{tab:corruptions} shows planning improvements across these conditions. Averaged over the six backbones evaluated under all four settings, L2 recovery is $57.6\%$, $26.0\%$, $24.7\%$, and $22.0\%$ for fog, snow, brightness, and physical perturbations, respectively; the corresponding collision recovery is $23.3\%$, $19.9\%$, $32.2\%$, and $30.4\%$. These averages use valid positive clean-to-noisy gaps. The results show benefits across diverse disturbances for most evaluated backbones. Full per-horizon L2 and collision results are reported in Appendix~\ref{app:b}.}

\begin{figure*}[!t]
  \centering
  \begin{subfigure}{\linewidth}
    \centering
    \begin{minipage}[t]{.49\linewidth}
      \vspace{0pt}
      \includegraphics[width=\linewidth,trim=20bp 53.5bp 492.05bp 223.21bp,clip]{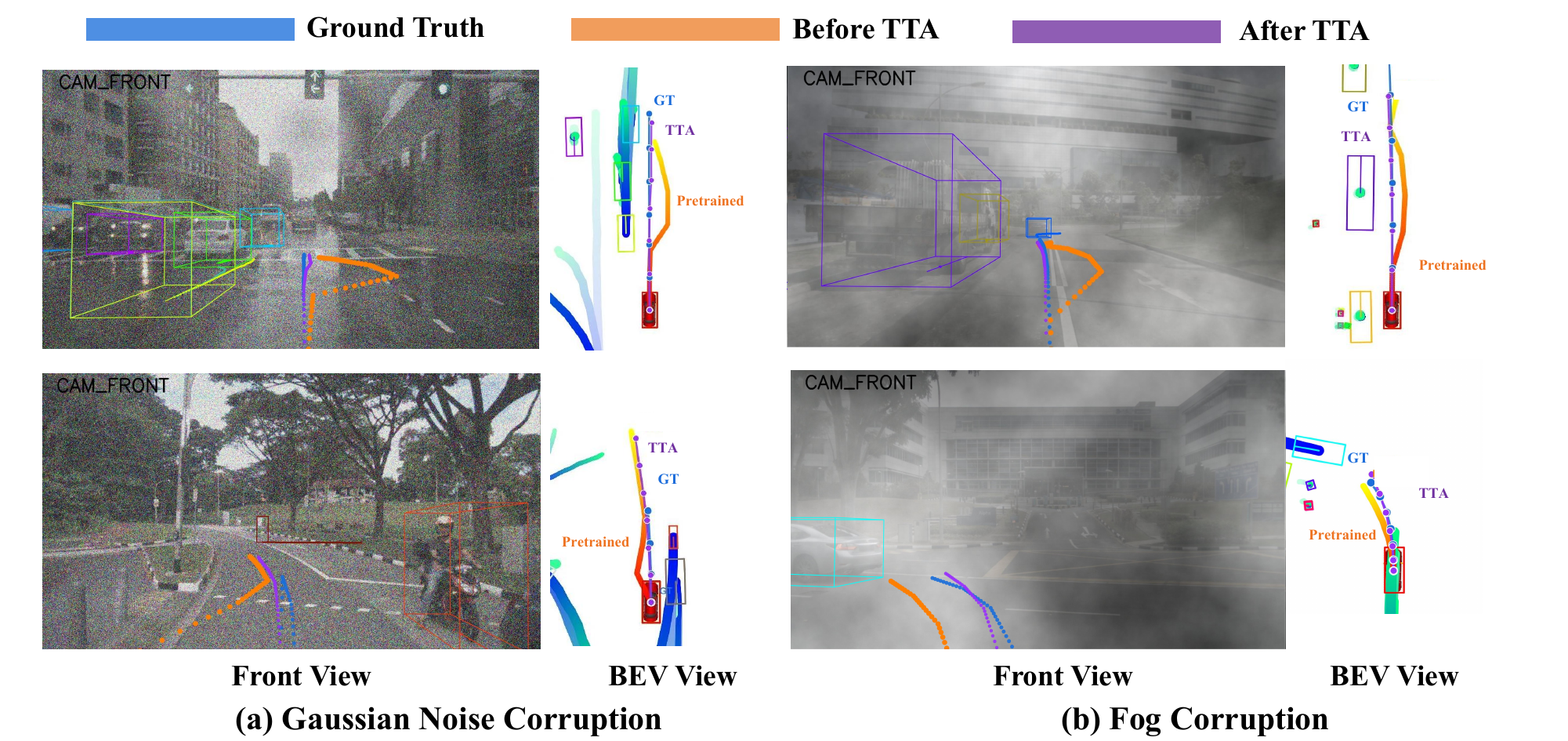}
    \end{minipage}\hfill
    \begin{minipage}[t]{.49\linewidth}
      \vspace{0pt}
      \includegraphics[width=\linewidth,trim=480bp 233.5bp 32.05bp 35bp,clip]{pdf/vis.pdf}
    \end{minipage}
    \caption{Trajectory comparisons under Gaussian noise (left) and fog (right).}
    \label{fig:vis-static}
  \end{subfigure}
  \par\smallskip
  \begin{subfigure}{\linewidth}
    \centering
    \includegraphics[width=\linewidth,trim=0bp 28.4bp 0bp 0bp,clip]{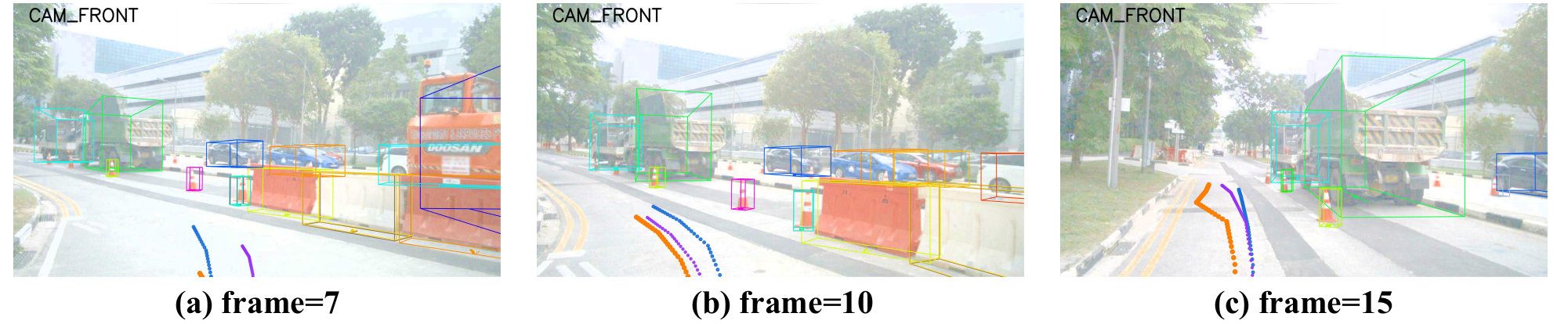}
    \par\vspace{1pt}
    \noindent\makebox[.3333\linewidth][c]{\footnotesize Frame~7}%
    \makebox[.3333\linewidth][c]{\footnotesize Frame~10}%
    \makebox[.3333\linewidth][c]{\footnotesize Frame~15}
    \caption{Trajectory evolution within one brightness-corrupted scene.}
    \label{fig:app-case-con}
  \end{subfigure}
  \caption{\textbf{Qualitative trajectory comparisons across scenarios.} Ground Truth (blue), Before TTA (orange), and After TTA (purple) show corrections under Gaussian noise and fog in (a), and progressive within-scene adaptation under brightness in (b).}
  \label{fig:vis}
\end{figure*}

{While the preceding results quantify overall planning performance, they do not directly illustrate how adaptation affects individual predicted trajectories. To provide this perspective, Figure~\ref{fig:vis}\subref{fig:vis-static} compares plans before and after TTA under severe Gaussian noise and fog. In the illustrated intersections and road-ending scenarios, the Before TTA plans exhibit abrupt bends, deviations from the road geometry, or paths directed toward surrounding vehicles. After adaptation, the predicted trajectories exhibit fewer abrupt deviations and align more closely with the ground-truth trajectories shown for comparison. Importantly, these ground-truth trajectories are used only for evaluation and visualization, rather than for test-time adaptation. \methodname{} relies solely on unlabeled test inputs and ego-state observations available up to the current time step. Figure~\ref{fig:vis}\subref{fig:app-case-con} further illustrates the gradual recovery of planning within a single scene under brightness corruption. As adaptation proceeds from frame~7 to~10 and then~15, the adapted trajectories become increasingly reasonable, exhibiting fewer abrupt deviations and better consistency with the local road geometry. Appendix~\ref{app:g} examines additional corrections and remaining failures.}

\begingroup
\color{black}
\begin{table}[t]
\caption{\textbf{NAVSIM performance under severity-5 Gaussian noise.} $\Delta$ denotes TTA $-$ Noisy in score points. NC, DAC, TTC, and EP denote no-at-fault collisions, drivable-area compliance, time-to-collision, and ego progress. PDMS reports the adapted score.}
\label{tab:navsim}
\centering
\small
\setlength{\tabcolsep}{3pt}
\resizebox{\linewidth}{!}{%
\begin{tabular}{@{}l|rrrr|c@{}}
\toprule
Planner & $\Delta$NC & $\Delta$DAC & $\Delta$TTC & $\Delta$EP & PDMS$\uparrow$ \\
\midrule
DiffusionDrive \citep{liao2025diffusiondrive} & +0.80 & +0.20 & +1.20 & +0.40 & \cellcolor{blue!5}\textbf{78.9}{\scriptsize (\textcolor{red}{+8.1\%})} \\
WoTE \citep{li2025wote} & +0.20 & +0.30 & +0.20 & +0.30 & \cellcolor{blue!5}\textbf{80.2}{\scriptsize (\textcolor{red}{+4.9\%})} \\
DriveLaW \citep{xia2026drivelaw} & +0.10 & +0.40 & +0.40 & +0.50 & \cellcolor{blue!5}\textbf{79.8}{\scriptsize (\textcolor{red}{+7.1\%})} \\
DriveVLA-W0 \citep{li2026drivevlaw0} & 0.00 & 0.00 & +0.60 & 0.00 & \cellcolor{blue!5}\textbf{86.4}{\scriptsize (\textcolor{red}{+30.0\%})} \\
GoalFlow \citep{xing2025goalflow} & +0.10 & +0.40 & +0.10 & +0.10 & \cellcolor{blue!5}\textbf{88.4}{\scriptsize (\textcolor{red}{+9.0\%})} \\
LTF \citep{beyonddrive} & +0.12 & +0.16 & +0.10 & +0.18 & \cellcolor{blue!5}\textbf{75.5}{\scriptsize (\textcolor{red}{+1.6\%})} \\
MeanFuser \citep{meanfuser} & +0.30 & +0.35 & +0.49 & +0.38 & \cellcolor{blue!5}\textbf{76.1}{\scriptsize (\textcolor{red}{+4.4\%})} \\
\bottomrule
\end{tabular}}
\end{table}

{The above nuScenes results primarily assess planning through displacement and collision metrics computed on recorded driving data. To assess the effectiveness of adapted trajectories during simulated execution, we therefore further evaluate \methodname{} on NAVSIM \citep{dauner2024navsim}. NAVSIM rolls out the model-predicted ego trajectory and evaluates its execution outcomes using the aggregated Predictive Driver Model Score (PDMS), which assesses safety, feasibility, comfort, and progress. Under severity-5 Gaussian noise, Table~\ref{tab:navsim} summarizes changes in collision avoidance, drivable-area compliance, time-to-collision, and ego progress relative to the noisy baselines, together with the adapted PDMS. PDMS improves across all seven planners, with an average recovery of $9.3\%$. Unlike nuScenes, where adaptation accumulates across consecutive frames within each scene, NAVSIM allows only a single adaptation step before generating the final trajectory for rollout. Complete clean, noisy, and adapted scores for NAVSIM metrics are provided in Appendix~\ref{app:a}. }

\par\endgroup

\begin{figure*}[!t]
  \centering
  \includegraphics[width=\linewidth]{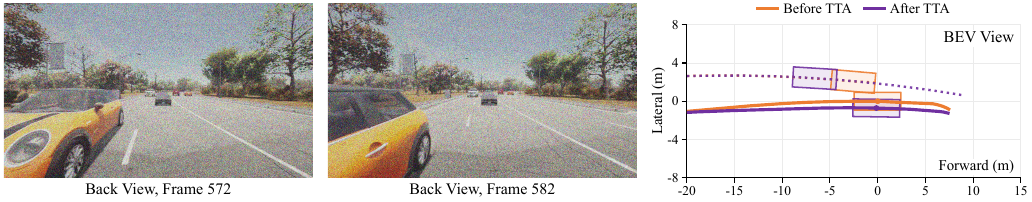}
  \caption{\textbf{Closed-loop cut-in case on Bench2Drive with UniAD.} Left and middle: rear-camera inputs from the noisy run. Right: a local BEV view of executed ego (solid) and cut-in vehicle (dotted) paths before TTA (orange) and after TTA (purple).}
  \label{fig:b2d-cutin-case}
\end{figure*}

To further evaluate \methodname{} in closed-loop driving, we present a UniAD case on Bench2Drive~\citep{jia2024bench2drive}. Under corrupted inputs, the unadapted model incurs a sideswipe collision during a cut-in interaction, whereas the adapted run completes the route without a recorded collision. Figure~\ref{fig:b2d-cutin-case} shows two rear-camera frames from the noisy run and the executed ego and cut-in vehicle paths from both runs in a shared BEV coordinate frame. Additional Bench2Drive results are provided in Appendix~\ref{app:b2d-results}.

\ifwidecomparisons\begin{table*}[!t]\else\begin{table}[!t]\fi
\caption{\textbf{{Cross-task adaptation and teacher--student component ablations.}} UniAD~\citep{hu2023planning} under severity-5 Gaussian noise.}
\label{tab:ablation}
\centering
\small
\setlength{\tabcolsep}{3pt}
\ifwidecomparisons\def\detailedtablewidth{0.52\linewidth}\else\def\detailedtablewidth{\linewidth}\fi
\sbox{\cvprtablebox}{%
\begin{tabular}{ccc|cccc|cccc}
\toprule
\multicolumn{3}{c|}{\textbf{Component}} & \multicolumn{4}{c|}{\textbf{L2 (m)$\downarrow$}} & \multicolumn{4}{c}{\textbf{Col. (\%)$\downarrow$}} \\
$\mathcal{L}_{mm}$ & $\mathcal{L}_{temp}$ & $\mathcal{S}_t$ & 1s & 2s & 3s & \textbf{AVG} & 1s & 2s & 3s & \textbf{AVG} \\
\midrule
- & - & - & 0.60 & 1.29 & 2.25 & \textbf{1.38} & 0.81 & 0.83 & 1.23 & \textbf{0.95} \\
\checkmark & - & - & 0.52 & 1.24 & 2.22 & \textbf{1.33}{\scriptsize (\textcolor{red}{+11.4\%})} & 0.70 & 0.91 & 1.15 & \textbf{0.92}{\scriptsize (\textcolor{red}{+4.5\%})} \\
- & \checkmark & - & 0.56 & 1.23 & 2.17 & \textbf{1.32}{\scriptsize (\textcolor{red}{+13.6\%})} & 0.81 & 0.80 & 1.15 & \textbf{0.92}{\scriptsize (\textcolor{red}{+4.5\%})} \\
\checkmark & \checkmark & - & 0.52 & 1.25 & 2.19 & \textbf{1.32}{\scriptsize (\textcolor{red}{+13.6\%})} & 0.43 & 0.85 & 1.16 & \textbf{0.81}{\scriptsize (\textcolor{red}{+21.2\%})} \\
\checkmark & - & \checkmark & 0.59 & 1.29 & 2.24 & \textbf{1.37}{\scriptsize (\textcolor{red}{+1.4\%})} & 0.85 & 0.53 & 1.10 & \textbf{0.83}{\scriptsize (\textcolor{red}{+18.9\%})} \\
- & \checkmark & \checkmark & 0.55 & 1.20 & 2.12 & \textbf{1.29}{\scriptsize (\textcolor{red}{+20.2\%})} & 0.80 & 0.85 & 1.33 & \textbf{0.99}{\scriptsize (\textcolor{red}{-6.3\%})} \\
\cellcolor{blue!5}\checkmark & \cellcolor{blue!5}\checkmark & \cellcolor{blue!5}\checkmark & \cellcolor{blue!5}0.51 & \cellcolor{blue!5}1.15 & \cellcolor{blue!5}2.09 & \cellcolor{blue!5}\textbf{1.25}{\scriptsize (\textcolor{red}{+29.5\%})} & \cellcolor{blue!5}0.53 & \cellcolor{blue!5}0.57 & \cellcolor{blue!5}1.08 & \cellcolor{blue!5}\textbf{0.72}{\scriptsize (\textcolor{red}{+34.8\%})} \\
\bottomrule
\end{tabular}}
\edef\ablationnaturalwidth{\the\wd\cvprtablebox}
\begin{subtable}{\detailedtablewidth}
\centering
\caption{{Joint objective and combined stabilization.}}
\label{tab:ablation-objectives}
\resizebox{\linewidth}{!}{\usebox{\cvprtablebox}}
\end{subtable}
\par\smallskip
\begin{subtable}{\detailedtablewidth}
\centering
\caption{{Individual teacher–student component ablations. EMA, Proj., and Acc. denote EMA anchoring, projection, and update acceptance/rollback, respectively.}}
\label{tab:ablation-stabilization}

\resizebox{\linewidth}{!}{%
\begin{tabular*}{\ablationnaturalwidth}{@{\extracolsep{\fill}}ccc|cccc|cccc}
\toprule
\multicolumn{3}{c|}{\textbf{Stabilization}} & \multicolumn{4}{c|}{\textbf{L2 (m)$\downarrow$}} & \multicolumn{4}{c}{\textbf{Col. (\%)$\downarrow$}} \\
EMA & Proj. & Acc. & 1s & 2s & 3s & \textbf{AVG} & 1s & 2s & 3s & \textbf{AVG} \\
\midrule
- & \checkmark & \checkmark & 0.56 & 1.23 & 2.17 & \textbf{1.32}{\scriptsize (\textcolor{red}{+12.8\%})} & 0.85 & 0.53 & 1.18 & \textbf{0.85}{\scriptsize (\textcolor{red}{+15.2\%})} \\
\checkmark & - & \checkmark & 0.56 & 1.21 & 2.15 & \textbf{1.31}{\scriptsize (\textcolor{red}{+15.2\%})} & 0.71 & 0.76 & 1.28 & \textbf{0.92}{\scriptsize (\textcolor{red}{+4.5\%})} \\
\checkmark & \checkmark & - & 0.54 & 1.19 & 2.11 & \textbf{1.28}{\scriptsize (\textcolor{red}{+21.5\%})} & 0.73 & 0.80 & 1.35 & \textbf{0.96}{\scriptsize (\textcolor{red}{-1.5\%})} \\
\bottomrule
\end{tabular*}}
\end{subtable}
\ifwidecomparisons\end{table*}\else\end{table}\fi

\subsection{{Ablation and Efficiency Analysis}}

{To validate the effectiveness of the proposed core modules, we conduct component ablations under severe Gaussian noise. Table~\ref{tab:ablation}\subref{tab:ablation-objectives} first evaluates cross-task adaptation on UniAD, progressively adding $\mathcal{L}_{mm}$, $\mathcal{L}_{temp}$, and the combined teacher--student mechanism $\mathcal{S}_t$.}
Perception entropy and temporal planning consistency each achieve $4.5\%$ collision recovery when used alone. Combining them increases collision recovery to $21.2\%$, while matching the $13.6\%$ L2 recovery of temporal consistency alone. Thus, the largest benefit of combining these two objectives in this ablation appears in collision-related performance.
With the same teacher–student stabilization enabled, our UniAD ablations in Table 5(a) reveal complementary trade-offs: temporal planning consistency improves trajectory accuracy while worsening collision performance, whereas perception entropy reduces collisions with little L2 recovery. Specifically, temporal consistency achieves $20.2\%$ L2 recovery but $-6.3\%$ collision recovery, while perception entropy achieves $18.9\%$ collision recovery with only $1.4\%$ L2 recovery. Coordinating both objectives achieves $29.5\%$ L2 recovery and $34.8\%$ collision recovery, increasing the latter by 13.6 percentage points over the joint objective without stabilization. Notably, the full model consistently reduces both L2 error and collision rate across all three future planning horizons (1s, 2s, and 3s) compared with the unadapted model, suggesting that the benefits of adaptation extend beyond retrospective ego-motion consistency to future trajectory planning. Together, these results support the need to coordinate perception and planning during adaptation to improve both trajectory accuracy and collision performance. Appendix~\ref{app:e} provides additional ablation results on other models.

{With both perception entropy and temporal planning consistency enabled, Table~\ref{tab:ablation}\subref{tab:ablation-stabilization} further examines the teacher–student mechanism by individually removing EMA anchoring, projection, or acceptance/rollback. All variants are evaluated against the same clean and noisy references as those in Table~\ref{tab:ablation}\subref{tab:ablation-objectives}. Removing these components reduces L2 recovery to 12.8\%, 15.2\%, and 21.5\%, respectively, compared with 29.5\% for the full method. Similarly, all three ablated variants exhibit inferior collision recovery, with the removal of acceptance/rollback even causing a slight degradation relative to the unadapted noisy model. These results demonstrate the complementary contributions of the three stabilization components to reliable test-time adaptation.}
\begin{figure}[!t]
  \centering
  \includegraphics[width=\linewidth]{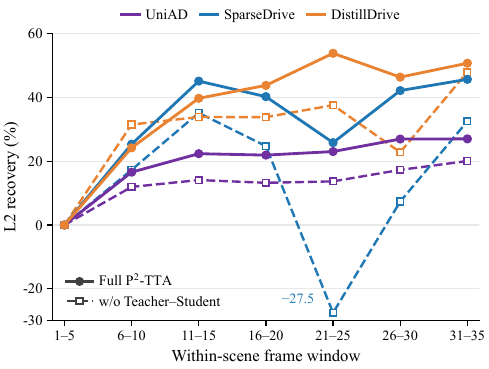}
  \caption{\textbf{Within-scene L2 recovery with and without teacher--student stabilization.} }
  \label{fig:long-horizon}
\end{figure}

To further evaluate the teacher--student mechanism during continual adaptation, Figure~\ref{fig:long-horizon} compares L2 recovery with and without this mechanism. Different frame windows cover different driving situations, leading to fluctuations in recovery rates. Despite these fluctuations, recovery generally increases as within-scene adaptation proceeds, while the teacher--student mechanism further improves recovery in most windows. Notably, on SparseDrive during frames~21--25, recovery drops to $-27.5\%$ without the mechanism, whereas the full method achieves $25.8\%$, illustrating its ability to mitigate local performance degradation. These results further support the effectiveness of the teacher--student mechanism during continual adaptation.

\ifwidecomparisons\begin{table*}[!t]\else\begin{table}[!t]\fi
\caption{\textcolor{black}{\textbf{Comparison with other TTA methods on UniAD~\citep{hu2023planning}.}} L2 and collision rate at each prediction horizon under severity-5 Gaussian noise. \underline{Underlining} marks the strongest baseline recovery for each metric.}
\label{tab:comparison}
\centering
\small
\setlength{\tabcolsep}{3pt}
\ifwidecomparisons\def\detailedtablewidth{0.52\linewidth}\else\def\detailedtablewidth{\linewidth}\fi
\resizebox{\detailedtablewidth}{!}{%
\begin{tabular}{@{}l|cccc|cccc@{}}
\toprule
\multirow{2}{*}{Method} & \multicolumn{4}{c|}{L2 (m)$\downarrow$} & \multicolumn{4}{c}{Col. (\%)$\downarrow$} \\
& 1s & 2s & 3s & AVG & 1s & 2s & 3s & AVG \\
\midrule
Without TTA & 0.60 & 1.29 & 2.25 & \textbf{1.38} & 0.81 & 0.83 & 1.23 & \textbf{0.95} \\
Tent \citep{wang2020tent} & 0.56 & 1.22 & 2.16 & \textbf{1.31}{\scriptsize (\textcolor{red}{+15.6\%})} & 0.85 & 0.65 & 1.15 & \textbf{0.88}{\scriptsize (\textcolor{red}{+10.5\%})} \\
EATA \citep{niu2022efficient} & 0.55 & 1.22 & 2.19 & \textbf{1.32}{\scriptsize (\textcolor{red}{+13.5\%})} & 0.55 & 0.52 & 1.15 & \textbf{0.74}{\scriptsize (\textcolor{red}{+31.8\%})} \\
EATA--E \citep{niu2022efficient} & 0.57 & 1.25 & 2.23 & \textbf{1.35}{\scriptsize (\textcolor{red}{+6.6\%})} & 0.55 & 0.52 & 1.13 & \underline{\textbf{0.73}{\scriptsize (\textcolor{red}{+32.8\%})}} \\
SAR \citep{niu2023towards} & 0.56 & 1.22 & 2.16 & \underline{\textbf{1.31}{\scriptsize (\textcolor{red}{+15.8\%})}} & 0.85 & 0.65 & 1.15 & \textbf{0.88}{\scriptsize (\textcolor{red}{+10.5\%})} \\
CoTTA \citep{wang2022continual} & 0.59 & 1.28 & 2.23 & \textbf{1.37}{\scriptsize (\textcolor{red}{+2.5\%})} & 0.81 & 0.52 & 1.13 & \textbf{0.82}{\scriptsize (\textcolor{red}{+19.8\%})} \\
\rowcolor{blue!5}{Ours} & 0.51 & 1.15 & 2.09 & \textbf{1.25}{\scriptsize (\textcolor{red}{+29.5\%})} & 0.53 & 0.57 & 1.08 & \textbf{0.72}{\scriptsize (\textcolor{red}{+34.8\%})} \\
\bottomrule
\end{tabular}}
\ifwidecomparisons\end{table*}\else\end{table}\fi

Table~\ref{tab:comparison} further compares \methodname{} with representative TTA baselines on UniAD under severe Gaussian noise using a shared adaptation objective. Our method achieves the highest recovery in both metrics, exceeding the respective strongest baselines by 13.7 and 2.0 percentage points in L2 and collision recovery, respectively. Together with the component ablations, these results support the effectiveness of our overall adaptation framework under the shared objective. Appendix~\ref{app:baseline-details} details the baseline implementations.

{{Finally, Table~\ref{tab:efficiency} evaluates the overhead of online adaptation.} {Across the six profiled models, \methodname{} updates an average of 0.11\textperthousand\ of parameters and requires 19.9\,ms per frame on average}, supporting recovery within a small additional computational budget. This lightweight update recovers $30.0\%$ of the collision-rate degradation on average. Among these models, UniAD has the highest adaptation cost and requires {$47.21$\,ms} per frame, supporting the practical use of {\methodname{}} across the evaluated architectures.}

\ifwidecomparisons\begin{table*}[!t]\else\begin{table}[!t]\fi
\caption{{\textbf{Parameter and TTA time efficiency.}
\methodname{} updates fewer than $0.23$\,\textperthousand of parameters and has a mean adaptation time below 50\,ms. }}
\centering
\small
\setlength{\tabcolsep}{3pt}
\ifwidecomparisons\def\efficiencytablewidth{0.48\linewidth}\else\def\efficiencytablewidth{\linewidth}\fi
\resizebox{\efficiencytablewidth}{!}{%
\begin{tabular}{@{}l|ccc|cc@{}}
\toprule
\multirow{2}{*}{\textbf{Model}} & \multicolumn{3}{c|}{\textbf{Parameters}} & \multicolumn{2}{c}{\textbf{Time (ms)}} \\
& {Updated} & {Total} & Ratio (\textperthousand) & Mean & Std \\
\midrule
UniAD \citep{hu2023planning} & {30.21K} & {131.82M} & 0.229 & {47.21} & {18.31} \\
ST-P3 \citep{hu2022st} & {1.18K} & {10.83M} & 0.109 & {13.70} & {8.10} \\
DiffusionDrive \citep{liao2025diffusiondrive} & {8.70K} & {92.04M} & 0.095 & {16.42} & {18.40} \\
SparseDrive \citep{sun2025sparsedrive} & {7.68K} & {86.14M} & 0.089 & {17.68} & 8.50 \\
DistillDrive \citep{yu2025distilldrive} & {7.68K} & {97.10M} & {0.079} & {16.37} & {13.12} \\
CoIRL-AD \citep{coirl} & {1.54K} & {51.45M} & {0.030} & {8.03} & {3.95} \\
\bottomrule
\end{tabular}}
\label{tab:efficiency}
\ifwidecomparisons\end{table*}\else\end{table}\fi

\section{Conclusion}
In this paper, we presented {\methodname{}}, a plug-and-play continual adaptation framework for end-to-end autonomous driving, designed for cross-task and cross-frame adaptation. {Its joint objective coordinates perception entropy minimization with temporal planning consistency, with ablations showing complementary gains in performance.} {\textcolor{black}{An EMA-centered projection and consistency-based acceptance mechanism constrain individual parameter updates.} } Extensive evaluations across diverse driving architectures demonstrate that {\methodname{}} effectively mitigates severe distribution shifts, achieving substantial reductions in displacement errors and collision rates {with low overhead.}

\clearpage
\section*{LLM Usage Statement}
OpenAI Codex was used to assist with manuscript editing, including language polishing, LaTeX formatting, and consistency checks. It was not used to generate experimental data or results. All AI-assisted content was reviewed and verified by the authors, who take responsibility for the final manuscript.

\clearpage
\setcounter{page}{1}
\renewcommand{\thepage}{S\arabic{page}}
\maketitlesupplementary
\noindent\textbf{Overview.} \hyperref[app:a]{Appendix~\ref*{app:a}} provides complete benchmark results and clean-input and time-varying-corruption evaluations; \hyperref[app:b]{Appendix~\ref*{app:b}} examines visual and physical robustness; \hyperref[app:c]{Appendix~\ref*{app:c}} evaluates geographic shifts and strong interactions; \hyperref[app:d]{Appendix~\ref*{app:d}} analyzes the loss weight; \hyperref[app:e]{Appendix~\ref*{app:e}} presents component ablations and within-scene continual adaptation; \hyperref[app:f]{Appendix~\ref*{app:f}} details model configurations and TTA baselines; \hyperref[app:g]{Appendix~\ref*{app:g}} analyzes qualitative cases and scenario-level errors; \hyperref[app:b2d-results]{Appendix~\ref*{app:b2d-results}} reports closed-loop Bench2Drive results; and \hyperref[app:h]{Appendix~\ref*{app:h}} discusses limitations and future work.

\appendix
\setcounter{topnumber}{5}
\setcounter{bottomnumber}{5}
\setcounter{totalnumber}{8}
\renewcommand{\topfraction}{0.95}
\renewcommand{\bottomfraction}{0.95}
\renewcommand{\textfraction}{0.05}
\renewcommand{\floatpagefraction}{0.85}
\setlength{\textfloatsep}{9pt plus 2pt minus 2pt}
\setlength{\floatsep}{8pt plus 2pt minus 2pt}
\setlength{\intextsep}{8pt plus 2pt minus 2pt}
\makeatletter
\setlength{\@fptop}{0pt}
\setlength{\@fpsep}{10pt}
\makeatother
\setcounter{table}{0}
\setcounter{figure}{0}
\renewcommand{\thetable}{S\arabic{table}}
\renewcommand{\thefigure}{S\arabic{figure}}
\renewcommand{\theHtable}{supp.\arabic{table}}
\renewcommand{\theHfigure}{supp.\arabic{figure}}
\section{\textcolor{black}{Detailed Benchmark Results and Robustness Evaluations}}\label{app:a}

\subsection{\textcolor{black}{Complete Benchmark Scores}}\label{app:main-details}
Tables~\ref{app:tab:gaussian-detailed} and~\ref{app:tab:navsim-detailed} expand the main nuScenes and NAVSIM results. Clean and Noisy denote unadapted scores; TTA denotes adaptation on the specified inputs. Blue shading marks TTA and red percentages denote recovery rates (RR).
For a lower-is-better metric, recovery is defined as
\begin{equation}
RR = \frac{M_{corrupted} - M_{adapted}}{M_{corrupted} - M_{clean}} \times 100\%.
\end{equation}
Here, $M_{clean}$, $M_{corrupted}$, and $M_{adapted}$ are the clean, noisy, and adapted scores. RR uses scores before display rounding and positive corruption-induced gaps; values above $100\%$ indicate performance exceeding the clean reference.

\begin{table}[!htbp]
\setlength{\tabcolsep}{1.5pt}
\caption{\textcolor{black}{\textbf{Complete nuScenes results under severity-5 Gaussian noise.} Clean is the unadapted reference; TTA uses corrupted inputs.}}
\label{app:tab:gaussian-detailed}
\centering\footnotesize
\setlength{\tabcolsep}{1.5pt}
\renewcommand{\arraystretch}{1.0}
\fitwidth{\linewidth}{%
\begin{tabular}{@{}l|cccc|cccc@{}}
\toprule
\multirow{2}{*}{Setting} & \multicolumn{4}{c|}{L2 (m)$\downarrow$} & \multicolumn{4}{c}{Col. (\%)$\downarrow$} \\
 & 1s & 2s & 3s & AVG & 1s & 2s & 3s & AVG \\
\midrule
\multicolumn{9}{@{}l}{\textbf{ST-P3~\citep{hu2022st}}} \\
Clean & 1.72 & 3.26 & 4.86 & \textbf{3.28} & 0.42 & 1.16 & 3.26 & \textbf{1.61} \\
Noisy & 2.51 & 5.22 & 8.13 & \textbf{5.29} & 0.89 & 5.08 & 9.42 & \textbf{5.13} \\
\rowcolor{blue!5}TTA & 2.29 & 4.57 & 6.95 & \textbf{4.60}{\scriptsize (\textcolor{red}{+34.3\%})} & 0.46 & 3.88 & 7.76 & \textbf{4.03}{\scriptsize (\textcolor{red}{+31.3\%})} \\
\specialrule{0.3pt}{0.3pt}{0.3pt}
\multicolumn{9}{@{}l}{\textbf{UniAD~\citep{hu2023planning}}} \\
Clean & 0.44 & 0.88 & 1.51 & \textbf{0.94} & 0.13 & 0.20 & 0.53 & \textbf{0.29} \\
Noisy & 0.60 & 1.29 & 2.25 & \textbf{1.38} & 0.81 & 0.83 & 1.23 & \textbf{0.95} \\
\rowcolor{blue!5}TTA & 0.51 & 1.15 & 2.09 & \textbf{1.25}{\scriptsize (\textcolor{red}{+29.5\%})} & 0.53 & 0.57 & 1.08 & \textbf{0.72}{\scriptsize (\textcolor{red}{+34.8\%})} \\
\specialrule{0.3pt}{0.3pt}{0.3pt}
\multicolumn{9}{@{}l}{\textbf{DistillDrive~\citep{yu2025distilldrive}}} \\
Clean & 0.28 & 0.55 & 0.91 & \textbf{0.58} & 0.00 & 0.03 & 0.18 & \textbf{0.07} \\
Noisy & 0.36 & 0.75 & 1.29 & \textbf{0.80} & 0.06 & 0.24 & 0.65 & \textbf{0.31} \\
\rowcolor{blue!5}TTA & 0.34 & 0.71 & 1.22 & \textbf{0.76}{\scriptsize (\textcolor{red}{+18.2\%})} & 0.06 & 0.20 & 0.57 & \textbf{0.28}{\scriptsize (\textcolor{red}{+12.5\%})} \\
\specialrule{0.3pt}{0.3pt}{0.3pt}
\multicolumn{9}{@{}l}{\textbf{SparseDrive~\citep{sun2025sparsedrive}}} \\
Clean & 0.40 & 1.03 & 1.95 & \textbf{1.13} & 0.02 & 0.12 & 0.76 & \textbf{0.30} \\
Noisy & 0.50 & 1.36 & 2.62 & \textbf{1.50} & 0.08 & 0.76 & 3.07 & \textbf{1.30} \\
\rowcolor{blue!5}TTA & 0.49 & 1.27 & 2.43 & \textbf{1.40}{\scriptsize (\textcolor{red}{+26.5\%})} & 0.10 & 0.49 & 1.95 & \textbf{0.85}{\scriptsize (\textcolor{red}{+45.5\%})} \\
\specialrule{0.3pt}{0.3pt}{0.3pt}
\multicolumn{9}{@{}l}{\textbf{DiffusionDrive~\citep{liao2025diffusiondrive}}} \\
Clean & 0.27 & 0.54 & 0.90 & \textbf{0.57} & 0.03 & 0.04 & 0.16 & \textbf{0.08} \\
Noisy & 0.34 & 0.72 & 1.25 & \textbf{0.77} & 0.02 & 0.18 & 0.68 & \textbf{0.29} \\
\rowcolor{blue!5}TTA & 0.32 & 0.68 & 1.21 & \textbf{0.74}{\scriptsize (\textcolor{red}{+15.0\%})} & 0.06 & 0.15 & 0.48 & \textbf{0.23}{\scriptsize (\textcolor{red}{+28.6\%})} \\
\specialrule{0.3pt}{0.3pt}{0.3pt}
\multicolumn{9}{@{}l}{\textbf{CoIRL-AD~\citep{coirl}}} \\
Clean & 0.29 & 0.59 & 1.00 & \textbf{0.63} & 0.06 & 0.10 & 0.37 & \textbf{0.18} \\
Noisy & 0.30 & 0.61 & 1.04 & \textbf{0.65} & 0.10 & 0.15 & 0.49 & \textbf{0.25} \\
\rowcolor{blue!5}TTA & 0.28 & 0.59 & 1.01 & \textbf{0.63}{\scriptsize (\textcolor{red}{+94.8\%})} & 0.10 & 0.15 & 0.44 & \textbf{0.23}{\scriptsize (\textcolor{red}{+27.3\%})} \\
\bottomrule
\end{tabular}}
\end{table}

\begin{table}[!htbp]
\caption{\textbf{NAVSIM results under severity-5 Gaussian noise.} Scores use a 0--100 scale. NC, DAC, TTC, and EP denote no-at-fault collisions, drivable-area compliance, time-to-collision, and ego progress; PDMS is the aggregate score. Shading marks TTA.}
\label{app:tab:navsim-detailed}
\centering
\small
\setlength{\tabcolsep}{2pt}
\renewcommand{\arraystretch}{1.0}
\fitwidth{\linewidth}{%
\begin{tabular}{@{}l|rrrrr@{}}
\toprule
Setting & NC$\uparrow$ & DAC$\uparrow$ & TTC$\uparrow$ & EP$\uparrow$ & PDMS$\uparrow$ \\
\midrule
\multicolumn{6}{@{}l}{\textbf{DiffusionDrive~\citep{liao2025diffusiondrive}}} \\
\textbf{Clean} & 99.2 & 96.1 & 94.8 & 81.8 & 88.0 \\
Noisy & 90.4 & 94.8 & 80.0 & 76.7 & 78.1 \\
\rowcolor{blue!5}TTA & \textbf{91.2}{\scriptsize (\textcolor{red}{+9.1\%})} & \textbf{95.0}{\scriptsize (\textcolor{red}{+16.0\%})} & \textbf{81.2}{\scriptsize (\textcolor{red}{+8.1\%})} & \textbf{77.1}{\scriptsize (\textcolor{red}{+7.8\%})} & \textbf{78.9}{\scriptsize (\textcolor{red}{+8.1\%})} \\
\specialrule{0.3pt}{0.3pt}{0.3pt}
\multicolumn{6}{@{}l}{\textbf{WoTE~\citep{li2025wote}}} \\
\textbf{Clean} & 99.4 & 96.2 & 96.5 & 82.3 & 89.9 \\
Noisy & 95.7 & 93.3 & 88.8 & 73.2 & 79.7 \\
\rowcolor{blue!5}TTA & \textbf{95.9}{\scriptsize (\textcolor{red}{+5.4\%})} & \textbf{93.6}{\scriptsize (\textcolor{red}{+10.4\%})} & \textbf{89.0}{\scriptsize (\textcolor{red}{+2.6\%})} & \textbf{73.5}{\scriptsize (\textcolor{red}{+3.3\%})} & \textbf{80.2}{\scriptsize (\textcolor{red}{+4.9\%})} \\
\specialrule{0.3pt}{0.3pt}{0.3pt}
\multicolumn{6}{@{}l}{\textbf{DriveLaW~\citep{xia2026drivelaw}}} \\
\textbf{Clean} & 99.0 & 95.0 & 96.5 & 79.8 & 89.0 \\
Noisy & 95.7 & 91.8 & 87.2 & 74.8 & 79.1 \\
\rowcolor{blue!5}TTA & \textbf{95.8}{\scriptsize (\textcolor{red}{+3.0\%})} & \textbf{92.2}{\scriptsize (\textcolor{red}{+12.5\%})} & \textbf{87.6}{\scriptsize (\textcolor{red}{+4.3\%})} & \textbf{75.3}{\scriptsize (\textcolor{red}{+10.0\%})} & \textbf{79.8}{\scriptsize (\textcolor{red}{+7.1\%})} \\
\specialrule{0.3pt}{0.3pt}{0.3pt}
\multicolumn{6}{@{}l}{\textbf{DriveVLA-W0~\citep{li2026drivevlaw0}}} \\
\textbf{Clean} & 98.4 & 95.3 & 96.2 & 80.9 & 87.1 \\
Noisy & 98.6 & 93.4 & 94.8 & 80.8 & 86.1 \\
\rowcolor{blue!5}TTA & \textbf{98.6} & \textbf{93.4}{\scriptsize (\textcolor{red}{0.0\%})} & \textbf{95.4}{\scriptsize (\textcolor{red}{+43.0\%})} & \textbf{80.8}{\scriptsize (\textcolor{red}{0.0\%})} & \textbf{86.4}{\scriptsize (\textcolor{red}{+30.0\%})} \\
\specialrule{0.3pt}{0.3pt}{0.3pt}
\multicolumn{6}{@{}l}{\textbf{GoalFlow~\citep{xing2025goalflow}}} \\
\textbf{Clean} & 98.4 & 98.3 & 93.9 & 85.2 & 91.4 \\
Noisy & 97.9 & 96.7 & 92.9 & 83.7 & 88.1 \\
\rowcolor{blue!5}TTA & \textbf{98.0}{\scriptsize (\textcolor{red}{+20.0\%})} & \textbf{97.1}{\scriptsize (\textcolor{red}{+25.0\%})} & \textbf{93.0}{\scriptsize (\textcolor{red}{+10.4\%})} & \textbf{83.8}{\scriptsize (\textcolor{red}{+6.9\%})} & \textbf{88.4}{\scriptsize (\textcolor{red}{+9.0\%})} \\
\specialrule{0.3pt}{0.3pt}{0.3pt}
\multicolumn{6}{@{}l}{\textbf{LTF~\citep{beyonddrive}}} \\
\textbf{Clean} & 98.29 & 97.08 & 94.60 & 83.01 & 88.77 \\
Noisy & 93.02 & 89.51 & 84.74 & 71.19 & 75.25 \\
\rowcolor{blue!5}TTA & \textbf{93.15}{\scriptsize (\textcolor{red}{+2.3\%})} & \textbf{89.68}{\scriptsize (\textcolor{red}{+2.2\%})} & \textbf{84.84}{\scriptsize (\textcolor{red}{+1.0\%})} & \textbf{71.37}{\scriptsize (\textcolor{red}{+1.5\%})} & \textbf{75.46}{\scriptsize (\textcolor{red}{+1.6\%})} \\
\specialrule{0.3pt}{0.3pt}{0.3pt}
\multicolumn{6}{@{}l}{\textbf{MeanFuser~\citep{meanfuser}}} \\
\textbf{Clean} & 98.26 & 97.41 & 94.46 & 83.29 & 89.02 \\
Noisy & 92.08 & 90.82 & 82.82 & 72.88 & 75.54 \\
\rowcolor{blue!5}TTA & \textbf{92.39}{\scriptsize (\textcolor{red}{+4.9\%})} & \textbf{91.17}{\scriptsize (\textcolor{red}{+5.2\%})} & \textbf{83.31}{\scriptsize (\textcolor{red}{+4.2\%})} & \textbf{73.25}{\scriptsize (\textcolor{red}{+3.6\%})} & \textbf{76.13}{\scriptsize (\textcolor{red}{+4.4\%})} \\
\bottomrule
\end{tabular}}
\end{table}

\subsection{\textcolor{black}{Adaptation on Clean Inputs}}

On clean inputs, adaptation reduces L2 and collision rate by $1.9\%$ and $3.4\%$ on average across five models (Table~\ref{app:tab:comparison}).

\begin{table}[!htbp]
\setlength{\tabcolsep}{1.5pt}
\caption{\textcolor{black}{\textbf{Adaptation on clean nuScenes inputs.} Both settings use the original images without synthetic corruption.}}
\label{app:tab:comparison}
\centering\footnotesize
\setlength{\tabcolsep}{1.5pt}
\renewcommand{\arraystretch}{1.02}
\fitwidth{\linewidth}{%
\begin{tabular}{@{}l|cccc|cccc@{}}
\toprule
\multirow{2}{*}{Setting} & \multicolumn{4}{c|}{L2 (m)$\downarrow$} & \multicolumn{4}{c}{Col. (\%)$\downarrow$} \\
 & 1s & 2s & 3s & AVG & 1s & 2s & 3s & AVG \\
\midrule
\multicolumn{9}{@{}l}{\textbf{ST-P3~\citep{hu2022st}}} \\
Before & 1.72 & 3.26 & 4.86 & \textbf{3.28} & 0.42 & 1.16 & 3.26 & \textbf{1.61} \\
\rowcolor{blue!5}TTA & 1.64 & 3.11 & 4.67 & \textbf{3.14} & 0.31 & 1.20 & 3.26 & \textbf{1.59} \\
\specialrule{0.3pt}{0.3pt}{0.3pt}
\multicolumn{9}{@{}l}{\textbf{UniAD~\citep{hu2023planning}}} \\
Before & 0.44 & 0.88 & 1.51 & \textbf{0.94} & 0.13 & 0.20 & 0.53 & \textbf{0.29} \\
\rowcolor{blue!5}TTA & 0.43 & 0.87 & 1.51 & \textbf{0.94} & 0.13 & 0.20 & 0.51 & \textbf{0.28} \\
\specialrule{0.3pt}{0.3pt}{0.3pt}
\multicolumn{9}{@{}l}{\textbf{DistillDrive~\citep{yu2025distilldrive}}} \\
Before & 0.28 & 0.55 & 0.91 & \textbf{0.58} & 0.00 & 0.03 & 0.18 & \textbf{0.07} \\
\rowcolor{blue!5}TTA & 0.28 & 0.54 & 0.90 & \textbf{0.57} & 0.00 & 0.03 & 0.18 & \textbf{0.07} \\
\specialrule{0.3pt}{0.3pt}{0.3pt}
\multicolumn{9}{@{}l}{\textbf{SparseDrive~\citep{sun2025sparsedrive}}} \\
Before & 0.31 & 0.60 & 0.99 & \textbf{0.63} & 0.01 & 0.05 & 0.18 & \textbf{0.08} \\
\rowcolor{blue!5}TTA & 0.31 & 0.58 & 0.98 & \textbf{0.62} & 0.00 & 0.04 & 0.20 & \textbf{0.08} \\
\specialrule{0.3pt}{0.3pt}{0.3pt}
\multicolumn{9}{@{}l}{\textbf{DiffusionDrive~\citep{liao2025diffusiondrive}}} \\
Before & 0.27 & 0.54 & 0.90 & \textbf{0.57} & 0.03 & 0.04 & 0.16 & \textbf{0.08} \\
\rowcolor{blue!5}TTA & 0.26 & 0.53 & 0.88 & \textbf{0.56} & 0.02 & 0.04 & 0.14 & \textbf{0.07} \\
\bottomrule
\end{tabular}}
\end{table}

\subsection{Temporally Evolving Corruption}
We increase contrast-corruption severity over 10 frames and decrease it over the next 10, retaining adapted parameters throughout. Table~\ref{app:tab:ood-temporal} shows mean L2 and collision reductions of $9.6\%$ and $15.5\%$ across five models, supporting adaptation under changing visual conditions.

\begin{table}[!htbp]
\setlength{\tabcolsep}{1.5pt}
\caption{\textcolor{black}{\textbf{Planning under temporally evolving contrast corruption.} Parameters persist across severity changes.}}
\label{app:tab:ood-temporal}
\centering\footnotesize
\setlength{\tabcolsep}{1.5pt}
\renewcommand{\arraystretch}{1.02}
\fitwidth{\linewidth}{%
\begin{tabular}{@{}l|cccc|cccc@{}}
\toprule
\multirow{2}{*}{Setting} & \multicolumn{4}{c|}{L2 (m)$\downarrow$} & \multicolumn{4}{c}{Col. (\%)$\downarrow$} \\
 & 1s & 2s & 3s & AVG & 1s & 2s & 3s & AVG \\
\midrule
\multicolumn{9}{@{}l}{\textbf{ST-P3~\citep{hu2022st}}} \\
Noisy & 3.47 & 6.72 & 10.28 & \textbf{6.82} & 0.41 & 4.94 & 9.47 & \textbf{4.94} \\
\rowcolor{blue!5}TTA & 2.76 & 5.58 & 8.60 & \textbf{5.65}{\scriptsize (\textcolor{red}{+33.2\%})} & 0.21 & 2.67 & 5.76 & \textbf{2.88}{\scriptsize (\textcolor{red}{+61.9\%})} \\
\specialrule{0.3pt}{0.3pt}{0.3pt}
\multicolumn{9}{@{}l}{\textbf{UniAD~\citep{hu2023planning}}} \\
Noisy & 0.55 & 1.15 & 2.01 & \textbf{1.24} & 0.56 & 0.48 & 0.78 & \textbf{0.61} \\
\rowcolor{blue!5}TTA & 0.50 & 1.05 & 1.87 & \textbf{1.14}{\scriptsize (\textcolor{red}{+32.6\%})} & 0.41 & 0.56 & 0.68 & \textbf{0.55}{\scriptsize (\textcolor{red}{+17.9\%})} \\
\specialrule{0.3pt}{0.3pt}{0.3pt}
\multicolumn{9}{@{}l}{\textbf{DistillDrive~\citep{yu2025distilldrive}}} \\
Noisy & 0.37 & 0.77 & 1.31 & \textbf{0.82} & 0.01 & 0.09 & 0.49 & \textbf{0.20} \\
\rowcolor{blue!5}TTA & 0.34 & 0.72 & 1.25 & \textbf{0.77}{\scriptsize (\textcolor{red}{+19.7\%})} & 0.02 & 0.10 & 0.43 & \textbf{0.18}{\scriptsize (\textcolor{red}{+10.5\%})} \\
\specialrule{0.3pt}{0.3pt}{0.3pt}
\multicolumn{9}{@{}l}{\textbf{SparseDrive~\citep{sun2025sparsedrive}}} \\
Noisy & 0.38 & 0.78 & 1.33 & \textbf{0.83} & 0.00 & 0.08 & 0.42 & \textbf{0.17} \\
\rowcolor{blue!5}TTA & 0.34 & 0.70 & 1.20 & \textbf{0.75}{\scriptsize (\textcolor{red}{+41.7\%})} & 0.02 & 0.09 & 0.36 & \textbf{0.16}{\scriptsize (\textcolor{red}{+11.5\%})} \\
\specialrule{0.3pt}{0.3pt}{0.3pt}
\multicolumn{9}{@{}l}{\textbf{DiffusionDrive~\citep{liao2025diffusiondrive}}} \\
Noisy & 0.38 & 0.78 & 1.33 & \textbf{0.83} & 0.01 & 0.09 & 0.49 & \textbf{0.20} \\
\rowcolor{blue!5}TTA & 0.34 & 0.73 & 1.25 & \textbf{0.77}{\scriptsize (\textcolor{red}{+21.8\%})} & 0.02 & 0.10 & 0.43 & \textbf{0.18}{\scriptsize (\textcolor{red}{+11.4\%})} \\
\bottomrule
\end{tabular}}
\end{table}

\section{\textcolor{black}{Robustness under Visual and Physical Perturbations}}\label{app:b}
Tables~\ref{app:tab:comparison-fog}--\ref{app:tab:joint-models} expand Table~\ref{tab:corruptions}. Cross-corruption averages use the six common backbones and positive clean-to-noisy gaps.

\subsection{Fog Corruption}
Under severity-5 fog, mean L2 and collision recovery reach $57.6\%$ and $23.3\%$ across six models (Table~\ref{app:tab:comparison-fog}).

\begin{table}[!htbp]
\setlength{\tabcolsep}{1.5pt}
\caption{\textcolor{black}{\textbf{Complete planning results under severity-5 fog corruption.}}}
\label{app:tab:comparison-fog}
\centering\footnotesize
\setlength{\tabcolsep}{1.5pt}
\renewcommand{\arraystretch}{1.02}
\fitwidth{\linewidth}{%
\begin{tabular}{@{}l|cccc|cccc@{}}
\toprule
\multirow{2}{*}{Setting} & \multicolumn{4}{c|}{L2 (m)$\downarrow$} & \multicolumn{4}{c}{Col. (\%)$\downarrow$} \\
 & 1s & 2s & 3s & AVG & 1s & 2s & 3s & AVG \\
\midrule
\multicolumn{9}{@{}l}{\textbf{ST-P3~\citep{hu2022st}}} \\
Noisy & 2.75 & 5.74 & 8.81 & \textbf{5.77} & 0.38 & 5.00 & 10.77 & \textbf{5.38} \\
\rowcolor{blue!5}TTA & 1.98 & 3.77 & 5.58 & \textbf{3.78}{\scriptsize (\textcolor{red}{+79.9\%})} & 0.64 & 2.70 & 6.12 & \textbf{3.15}{\scriptsize (\textcolor{red}{+59.2\%})} \\
\specialrule{0.3pt}{0.3pt}{0.3pt}
\multicolumn{9}{@{}l}{\textbf{UniAD~\citep{hu2023planning}}} \\
Noisy & 0.47 & 0.96 & 1.66 & \textbf{1.03} & 0.17 & 0.32 & 0.47 & \textbf{0.32} \\
\rowcolor{blue!5}TTA & 0.45 & 0.92 & 1.62 & \textbf{1.00}{\scriptsize (\textcolor{red}{+33.3\%})} & 0.17 & 0.31 & 0.45 & \textbf{0.31}{\scriptsize (\textcolor{red}{+33.3\%})} \\
\specialrule{0.3pt}{0.3pt}{0.3pt}
\multicolumn{9}{@{}l}{\textbf{DistillDrive~\citep{yu2025distilldrive}}} \\
Noisy & 0.30 & 0.60 & 0.99 & \textbf{0.63} & 0.03 & 0.08 & 0.25 & \textbf{0.12} \\
\rowcolor{blue!5}TTA & 0.29 & 0.58 & 0.97 & \textbf{0.61}{\scriptsize (\textcolor{red}{+40.0\%})} & 0.02 & 0.07 & 0.25 & \textbf{0.11}{\scriptsize (\textcolor{red}{+20.0\%})} \\
\specialrule{0.3pt}{0.3pt}{0.3pt}
\multicolumn{9}{@{}l}{\textbf{SparseDrive~\citep{sun2025sparsedrive}}} \\
Noisy & 0.31 & 0.60 & 0.98 & \textbf{0.63} & 0.00 & 0.07 & 0.28 & \textbf{0.12} \\
\rowcolor{blue!5}TTA & 0.30 & 0.59 & 0.97 & \textbf{0.62}{\scriptsize (\textcolor{red}{+21.1\%})} & 0.02 & 0.06 & 0.26 & \textbf{0.11}{\scriptsize (\textcolor{red}{+25.0\%})} \\
\specialrule{0.3pt}{0.3pt}{0.3pt}
\multicolumn{9}{@{}l}{\textbf{DiffusionDrive~\citep{liao2025diffusiondrive}}} \\
Noisy & 0.29 & 0.58 & 0.98 & \textbf{0.62} & 0.02 & 0.09 & 0.25 & \textbf{0.12} \\
\rowcolor{blue!5}TTA & 0.27 & 0.55 & 0.92 & \textbf{0.58}{\scriptsize (\textcolor{red}{+80.0\%})} & 0.02 & 0.09 & 0.24 & \textbf{0.12}{\scriptsize (\textcolor{red}{0.0\%})} \\
\specialrule{0.3pt}{0.3pt}{0.3pt}
\multicolumn{9}{@{}l}{\textbf{CoIRL-AD~\citep{coirl}}} \\
Noisy & 0.30 & 0.60 & 1.02 & \textbf{0.64} & 0.30 & 0.26 & 0.51 & \textbf{0.36} \\
\rowcolor{blue!5}TTA & 0.29 & 0.59 & 1.00 & \textbf{0.63}{\scriptsize (\textcolor{red}{+91.1\%})} & 0.29 & 0.25 & 0.51 & \textbf{0.35}{\scriptsize (\textcolor{red}{+2.4\%})} \\
\bottomrule
\end{tabular}}
\end{table}

\subsection{Snow Corruption}
Snow causes the largest average degradation among the three visual corruptions. Adaptation recovers $26.0\%$ of L2 and $19.9\%$ of collision degradation (Table~\ref{app:tab:snow}).

\begin{table}[!htbp]
\setlength{\tabcolsep}{1.5pt}
\caption{\textcolor{black}{\textbf{Complete planning results under snow corruption.}}}
\label{app:tab:snow}
\centering\footnotesize
\setlength{\tabcolsep}{1.5pt}
\renewcommand{\arraystretch}{1.02}
\fitwidth{\linewidth}{%
\begin{tabular}{@{}l|cccc|cccc@{}}
\toprule
\multirow{2}{*}{Setting} & \multicolumn{4}{c|}{L2 (m)$\downarrow$} & \multicolumn{4}{c}{Col. (\%)$\downarrow$} \\
 & 1s & 2s & 3s & AVG & 1s & 2s & 3s & AVG \\
\midrule
\multicolumn{9}{@{}l}{\textbf{ST-P3~\citep{hu2022st}}} \\
Noisy & 3.67 & 7.67 & 11.72 & \textbf{7.69} & 2.88 & 8.23 & 8.64 & \textbf{6.58} \\
\rowcolor{blue!5}TTA & 2.97 & 6.54 & 10.12 & \textbf{6.54}{\scriptsize (\textcolor{red}{+26.1\%})} & 1.03 & 3.09 & 6.17 & \textbf{3.43}{\scriptsize (\textcolor{red}{+63.4\%})} \\
\specialrule{0.3pt}{0.3pt}{0.3pt}
\multicolumn{9}{@{}l}{\textbf{UniAD~\citep{hu2023planning}}} \\
Noisy & 0.44 & 1.10 & 2.00 & \textbf{1.18} & 0.38 & 1.13 & 1.41 & \textbf{0.97} \\
\rowcolor{blue!5}TTA & 0.42 & 1.04 & 1.93 & \textbf{1.13}{\scriptsize (\textcolor{red}{+20.8\%})} & 0.48 & 0.68 & 1.34 & \textbf{0.84}{\scriptsize (\textcolor{red}{+19.1\%})} \\
\specialrule{0.3pt}{0.3pt}{0.3pt}
\multicolumn{9}{@{}l}{\textbf{DistillDrive~\citep{yu2025distilldrive}}} \\
Noisy & 0.46 & 0.92 & 1.53 & \textbf{0.97} & 0.19 & 0.53 & 0.58 & \textbf{0.44} \\
\rowcolor{blue!5}TTA & 0.40 & 0.82 & 1.38 & \textbf{0.87}{\scriptsize (\textcolor{red}{+25.6\%})} & 0.19 & 0.48 & 0.58 & \textbf{0.42}{\scriptsize (\textcolor{red}{+5.4\%})} \\
\specialrule{0.3pt}{0.3pt}{0.3pt}
\multicolumn{9}{@{}l}{\textbf{SparseDrive~\citep{sun2025sparsedrive}}} \\
Noisy & 0.42 & 0.81 & 1.33 & \textbf{0.85} & 0.19 & 0.44 & 0.84 & \textbf{0.49} \\
\rowcolor{blue!5}TTA & 0.38 & 0.75 & 1.24 & \textbf{0.80}{\scriptsize (\textcolor{red}{+22.7\%})} & 0.09 & 0.43 & 0.90 & \textbf{0.48}{\scriptsize (\textcolor{red}{+2.4\%})} \\
\specialrule{0.3pt}{0.3pt}{0.3pt}
\multicolumn{9}{@{}l}{\textbf{DiffusionDrive~\citep{liao2025diffusiondrive}}} \\
Noisy & 0.47 & 0.98 & 1.65 & \textbf{1.03} & 0.04 & 0.22 & 0.95 & \textbf{0.40} \\
\rowcolor{blue!5}TTA & 0.42 & 0.88 & 1.51 & \textbf{0.94}{\scriptsize (\textcolor{red}{+19.6\%})} & 0.07 & 0.23 & 0.80 & \textbf{0.37}{\scriptsize (\textcolor{red}{+9.4\%})} \\
\specialrule{0.3pt}{0.3pt}{0.3pt}
\multicolumn{9}{@{}l}{\textbf{CoIRL-AD~\citep{coirl}}} \\
Noisy & 0.30 & 0.61 & 1.03 & \textbf{0.64} & 0.35 & 0.43 & 0.81 & \textbf{0.53} \\
\rowcolor{blue!5}TTA & 0.29 & 0.60 & 1.02 & \textbf{0.64}{\scriptsize (\textcolor{red}{+41.0\%})} & 0.27 & 0.38 & 0.74 & \textbf{0.46}{\scriptsize (\textcolor{red}{+19.6\%})} \\
\bottomrule
\end{tabular}}
\end{table}

\subsection{Brightness Corruption}
Under brightness changes, mean L2 and collision recovery reach $24.7\%$ and $32.2\%$ (Table~\ref{app:tab:brightness}), extending the gains to illumination shifts.

\begin{table}[!htbp]
\setlength{\tabcolsep}{1.5pt}
\caption{\textcolor{black}{\textbf{Complete planning results under brightness corruption.}}}
\label{app:tab:brightness}
\centering\footnotesize
\setlength{\tabcolsep}{1.5pt}
\renewcommand{\arraystretch}{1.02}
\fitwidth{\linewidth}{%
\begin{tabular}{@{}l|cccc|cccc@{}}
\toprule
\multirow{2}{*}{Setting} & \multicolumn{4}{c|}{L2 (m)$\downarrow$} & \multicolumn{4}{c}{Col. (\%)$\downarrow$} \\
 & 1s & 2s & 3s & AVG & 1s & 2s & 3s & AVG \\
\midrule
\multicolumn{9}{@{}l}{\textbf{ST-P3~\citep{hu2022st}}} \\
Noisy & 2.10 & 4.00 & 6.05 & \textbf{4.06} & 0.62 & 2.01 & 4.63 & \textbf{2.42} \\
\rowcolor{blue!5}TTA & 1.95 & 3.73 & 5.61 & \textbf{3.76}{\scriptsize (\textcolor{red}{+38.5\%})} & 0.58 & 1.54 & 4.23 & \textbf{2.11}{\scriptsize (\textcolor{red}{+38.3\%})} \\
\specialrule{0.3pt}{0.3pt}{0.3pt}
\multicolumn{9}{@{}l}{\textbf{UniAD~\citep{hu2023planning}}} \\
Noisy & 0.46 & 0.91 & 1.57 & \textbf{0.98} & 0.23 & 0.30 & 0.45 & \textbf{0.33} \\
\rowcolor{blue!5}TTA & 0.44 & 0.89 & 1.55 & \textbf{0.97}{\scriptsize (\textcolor{red}{+34.0\%})} & 0.24 & 0.28 & 0.44 & \textbf{0.32}{\scriptsize (\textcolor{red}{+18.2\%})} \\
\specialrule{0.3pt}{0.3pt}{0.3pt}
\multicolumn{9}{@{}l}{\textbf{DistillDrive~\citep{yu2025distilldrive}}} \\
Noisy & 0.28 & 0.56 & 0.95 & \textbf{0.60} & 0.04 & 0.09 & 0.23 & \textbf{0.12} \\
\rowcolor{blue!5}TTA & 0.27 & 0.56 & 0.95 & \textbf{0.60}{\scriptsize (\textcolor{red}{+21.6\%})} & 0.00 & 0.05 & 0.21 & \textbf{0.09}{\scriptsize (\textcolor{red}{+62.5\%})} \\
\specialrule{0.3pt}{0.3pt}{0.3pt}
\multicolumn{9}{@{}l}{\textbf{SparseDrive~\citep{sun2025sparsedrive}}} \\
Noisy & 0.39 & 0.77 & 1.30 & \textbf{0.82} & 0.00 & 0.20 & 0.42 & \textbf{0.20} \\
\rowcolor{blue!5}TTA & 0.35 & 0.70 & 1.19 & \textbf{0.75}{\scriptsize (\textcolor{red}{+36.6\%})} & 0.00 & 0.18 & 0.36 & \textbf{0.18}{\scriptsize (\textcolor{red}{+16.5\%})} \\
\specialrule{0.3pt}{0.3pt}{0.3pt}
\multicolumn{9}{@{}l}{\textbf{DiffusionDrive~\citep{liao2025diffusiondrive}}} \\
Noisy & 0.36 & 0.71 & 1.20 & \textbf{0.75} & 0.00 & 0.18 & 0.42 & \textbf{0.20} \\
\rowcolor{blue!5}TTA & 0.34 & 0.68 & 1.15 & \textbf{0.72}{\scriptsize (\textcolor{red}{+16.3\%})} & 0.00 & 0.09 & 0.30 & \textbf{0.13}{\scriptsize (\textcolor{red}{+57.9\%})} \\
\specialrule{0.3pt}{0.3pt}{0.3pt}
\multicolumn{9}{@{}l}{\textbf{CoIRL-AD~\citep{coirl}}} \\
Noisy & 0.31 & 0.62 & 1.04 & \textbf{0.66} & 0.15 & 0.17 & 0.47 & \textbf{0.26} \\
\rowcolor{blue!5}TTA & 0.31 & 0.62 & 1.04 & \textbf{0.66}{\scriptsize (\textcolor{red}{+1.3\%})} & 0.15 & 0.17 & 0.47 & \textbf{0.26}{\scriptsize (\textcolor{red}{0.0\%})} \\
\bottomrule
\end{tabular}}
\end{table}

\subsection{\textcolor{black}{Concurrent Perturbation Severity on UniAD}}
We combine GPS and yaw noise with motion blur to test adaptation when both visual inputs and observed-motion feedback are corrupted.

GPS and yaw magnitudes are standard deviations of zero-mean Gaussian errors. Perturbations are keyed by sample token and seed, matching the noisy baseline and TTA. Consecutive noisy positions are differenced and transformed into the previous planning frame using the noisy orientation and native axis conventions.

Noise enters UniAD through \texttt{can\_bus}, and SparseDrive, DistillDrive, and nuScenes DiffusionDrive through lidar-to-global poses. ST-P3 uses corrupted ego-motion inputs and displacement references. These perturbations affect state inputs and temporal feedback; evaluation trajectories, object annotations, label-construction poses, and camera calibration remain clean.

On UniAD, GPS standard deviations of $0.2$, $0.5$, $1$, $2$, and $4$\,m are paired with yaw deviations of $0.5^{\circ}$, $1^{\circ}$, $2^{\circ}$, $5^{\circ}$, and $10^{\circ}$; motion blur uses \texttt{imagecorruptions}. Table~\ref{app:tab:joint-severity} shows improvement in both metrics at severities 1--3, with $57.5\%$--$72.1\%$ L2 recovery.

L2 recovery drops to $3.9\%$ and $-0.3\%$ at severities 4 and 5, respectively, while collision recovery remains $52.4\%$ at severity 5. Severe localization errors therefore limit the usefulness of motion feedback for trajectory accuracy.

\begin{table}[!htbp]
\setlength{\tabcolsep}{1.5pt}
\caption{\textbf{UniAD under concurrent localization noise and motion blur.}}
\centering
\fitwidth{\linewidth}{
\small
{\color{black}
\begin{tabular}{l|cccc|cccc}
\toprule
\multirow{2}{*}{\textbf{Severity (GPS/yaw)}} & \multicolumn{4}{c|}{\textbf{L2 (m)$\downarrow$}} & \multicolumn{4}{c}{\textbf{Col. (\%)$\downarrow$}} \\
& 1s & 2s & 3s & \textbf{AVG} & 1s & 2s & 3s & \textbf{AVG} \\
\midrule
1 (0.2 m/$0.5^{\circ}$) & 0.77 & 1.53 & 2.55 & \textbf{1.62} & 0.47 & 0.19 & 1.89 & \textbf{0.85} \\
2 (0.5 m/$1^{\circ}$) & 0.94 & 1.77 & 2.83 & \textbf{1.85} & 0.68 & 0.24 & 2.74 & \textbf{1.22} \\
3 (1 m/$2^{\circ}$) & 0.92 & 1.63 & 2.48 & \textbf{1.68} & 0.70 & 0.24 & 2.96 & \textbf{1.30} \\
4 (2 m/$5^{\circ}$) & 1.35 & 2.31 & 3.24 & \textbf{2.30} & 0.98 & 0.43 & 3.84 & \textbf{1.75} \\
5 (4 m/$10^{\circ}$) & 2.43 & 4.16 & 5.57 & \textbf{4.05} & 1.73 & 1.17 & 5.15 & \textbf{2.68} \\
\midrule
\cellcolor{blue!5}1 (0.2 m/$0.5^{\circ}$) & \cellcolor{blue!5}0.50 & \cellcolor{blue!5}1.04 & \cellcolor{blue!5}1.88 & \cellcolor{blue!5}\textbf{1.14}{\scriptsize (\textcolor{red}{+70.4\%})} & \cellcolor{blue!5}0.13 & \cellcolor{blue!5}0.07 & \cellcolor{blue!5}0.50 & \cellcolor{blue!5}\textbf{0.23}{\scriptsize (\textcolor{red}{+110.1\%})} \\
\cellcolor{blue!5}2 (0.5 m/$1^{\circ}$) & \cellcolor{blue!5}0.57 & \cellcolor{blue!5}1.10 & \cellcolor{blue!5}1.91 & \cellcolor{blue!5}\textbf{1.19}{\scriptsize (\textcolor{red}{+72.1\%})} & \cellcolor{blue!5}0.15 & \cellcolor{blue!5}0.08 & \cellcolor{blue!5}0.64 & \cellcolor{blue!5}\textbf{0.29}{\scriptsize (\textcolor{red}{+100.0\%})} \\
\cellcolor{blue!5}3 (1 m/$2^{\circ}$) & \cellcolor{blue!5}0.77 & \cellcolor{blue!5}1.37 & \cellcolor{blue!5}1.62 & \cellcolor{blue!5}\textbf{1.25}{\scriptsize (\textcolor{red}{+57.5\%})} & \cellcolor{blue!5}0.23 & \cellcolor{blue!5}0.07 & \cellcolor{blue!5}0.65 & \cellcolor{blue!5}\textbf{0.32}{\scriptsize (\textcolor{red}{+97.4\%})} \\
\cellcolor{blue!5}4 (2 m/$5^{\circ}$) & \cellcolor{blue!5}1.33 & \cellcolor{blue!5}2.25 & \cellcolor{blue!5}3.16 & \cellcolor{blue!5}\textbf{2.25}{\scriptsize (\textcolor{red}{+3.9\%})} & \cellcolor{blue!5}0.70 & \cellcolor{blue!5}0.23 & \cellcolor{blue!5}3.37 & \cellcolor{blue!5}\textbf{1.43}{\scriptsize (\textcolor{red}{+21.7\%})} \\
\cellcolor{blue!5}5 (4 m/$10^{\circ}$) & \cellcolor{blue!5}2.44 & \cellcolor{blue!5}4.17 & \cellcolor{blue!5}5.58 & \cellcolor{blue!5}\textbf{4.06}{\scriptsize (\textcolor{red}{-0.3\%})} & \cellcolor{blue!5}1.36 & \cellcolor{blue!5}0.83 & \cellcolor{blue!5}2.10 & \cellcolor{blue!5}\textbf{1.43}{\scriptsize (\textcolor{red}{+52.4\%})} \\
\bottomrule
\end{tabular}
}
}
\label{app:tab:joint-severity}
\end{table}

\subsection{\textcolor{black}{Compound Perturbations across Architectures}}
At severity 3 ($1$\,m GPS noise, $2^{\circ}$ yaw noise, and motion blur), six architectures achieve mean L2 and collision recovery of $22.0\%$ and $30.4\%$ over positive corruption-induced gaps (Table~\ref{app:tab:joint-models}).

\begin{table}[!htbp]
\setlength{\tabcolsep}{1.5pt}
\caption{\textbf{Concurrent localization and visual perturbations at severity 3.}}
\centering
\fitwidth{\linewidth}{
\small
{\color{black}
\begin{tabular}{l|cccc|cccc}
\toprule
\multirow{2}{*}{\textbf{Model}} & \multicolumn{4}{c|}{\textbf{L2 (m)$\downarrow$}} & \multicolumn{4}{c}{\textbf{Col. (\%)$\downarrow$}} \\
& 1s & 2s & 3s & \textbf{AVG} & 1s & 2s & 3s & \textbf{AVG} \\
\midrule
ST-P3 \citep{hu2022st} & 2.00 & 3.90 & 5.93 & \textbf{3.94} & 0.79 & 3.17 & 5.27 & \textbf{3.08} \\
UniAD \citep{hu2023planning} & 0.92 & 1.63 & 2.48 & \textbf{1.68} & 0.70 & 0.24 & 2.96 & \textbf{1.30} \\
DistillDrive \citep{yu2025distilldrive} & 0.62 & 1.11 & 1.70 & \textbf{1.14} & 0.20 & 0.68 & 1.42 & \textbf{0.77} \\
SparseDrive \citep{sun2025sparsedrive} &  0.69 & 1.23 & 1.87 & \textbf{1.26} & 0.19 & 0.63 & 1.45 & \textbf{0.76} \\
DiffusionDrive \citep{liao2025diffusiondrive} & 0.54 & 0.98 &1.49  & \textbf{1.00} & 0.00 & 0.24 & 0.55 & \textbf{0.26} \\
CoIRL-AD \citep{coirl} & 0.28 & 0.59 & 1.01 & \textbf{0.63} & 0.07 & 0.11 & 0.39 & \textbf{0.19} \\
\midrule
\cellcolor{blue!5}ST-P3 \citep{hu2022st} & \cellcolor{blue!5}1.93 & \cellcolor{blue!5}3.78 & \cellcolor{blue!5}5.75 & \cellcolor{blue!5}\textbf{3.82}{\scriptsize (\textcolor{red}{+18.6\%})} & \cellcolor{blue!5}0.60 & \cellcolor{blue!5}2.39 & \cellcolor{blue!5}4.07 & \cellcolor{blue!5}\textbf{2.35}{\scriptsize (\textcolor{red}{+49.3\%})} \\

\cellcolor{blue!5}UniAD \citep{hu2023planning} & \cellcolor{blue!5}0.77 & \cellcolor{blue!5}1.37 & \cellcolor{blue!5}1.62 & \cellcolor{blue!5}\textbf{1.25}{\scriptsize (\textcolor{red}{+57.5\%})} & \cellcolor{blue!5}0.23 & \cellcolor{blue!5}0.07 & \cellcolor{blue!5}0.65 & \cellcolor{blue!5}\textbf{0.32}{\scriptsize (\textcolor{red}{+97.4\%})} \\

\cellcolor{blue!5}DistillDrive \citep{yu2025distilldrive} & \cellcolor{blue!5}0.61 & \cellcolor{blue!5}1.10 & \cellcolor{blue!5}1.68 & \cellcolor{blue!5}\textbf{1.13}{\scriptsize (\textcolor{red}{+2.4\%})} & \cellcolor{blue!5}0.15 & \cellcolor{blue!5}0.61 & \cellcolor{blue!5}1.35 & \cellcolor{blue!5}\textbf{0.70}{\scriptsize (\textcolor{red}{+9.1\%})} \\

\cellcolor{blue!5}SparseDrive \citep{sun2025sparsedrive} & \cellcolor{blue!5}0.66 & \cellcolor{blue!5}1.16 & \cellcolor{blue!5}1.80 & \cellcolor{blue!5}\textbf{1.21}{\scriptsize (\textcolor{red}{+8.9\%})} & \cellcolor{blue!5}0.00 & \cellcolor{blue!5}0.58 & \cellcolor{blue!5}1.26 & \cellcolor{blue!5}\textbf{0.61}{\scriptsize (\textcolor{red}{+21.2\%})} \\

\cellcolor{blue!5}DiffusionDrive \citep{liao2025diffusiondrive} & \cellcolor{blue!5}0.53 & \cellcolor{blue!5}0.97 & \cellcolor{blue!5}1.48 & \cellcolor{blue!5}\textbf{0.99}{\scriptsize (\textcolor{red}{+2.3\%})} & \cellcolor{blue!5}0.00 & \cellcolor{blue!5}0.24 & \cellcolor{blue!5}0.52 & \cellcolor{blue!5}\textbf{0.25}{\scriptsize (\textcolor{red}{+5.5\%})} \\
\cellcolor{blue!5}CoIRL-AD \citep{coirl} & \cellcolor{blue!5}0.28 & \cellcolor{blue!5}0.58 & \cellcolor{blue!5}1.01 & \cellcolor{blue!5}\textbf{0.63}{\scriptsize (\textcolor{red}{+42.5\%})} & \cellcolor{blue!5}0.07 & \cellcolor{blue!5}0.11 & \cellcolor{blue!5}0.39 & \cellcolor{blue!5}\textbf{0.19}{\scriptsize (\textcolor{red}{0.0\%})} \\
\bottomrule
\end{tabular}
}
}
\label{app:tab:joint-models}
\end{table}

\section{Generalization under Out-of-Distribution Conditions}\label{app:c}

\subsection{Cross-City Transfer}
We train SparseDrive exclusively on Boston data, which accounts for $55.8\%$ of the nuScenes training set, using 100 epochs for Stage~1 and 10 epochs for Stage~2. We then evaluate the model in Holland Village, One-North, and Queenstown in Singapore without synthetic corruption. Boston serves as the source-domain reference in Table~\ref{tab:cross-city}.

Both metrics improve in all three target towns, with $9.9\%$--$35.6\%$ L2 recovery and $8.0\%$--$45.5\%$ collision recovery relative to Boston (Table~\ref{tab:cross-city}).

\begin{table}[!htbp]
\setlength{\tabcolsep}{1.5pt}

\caption{{\textbf{Cross-city transfer on SparseDrive.} Results use SparseDrive~\citep{sun2025sparsedrive}; RR takes the unadapted Boston scores as its reference.}}
\label{tab:cross-city}
\centering
\small
\setlength{\tabcolsep}{3pt}
\fitwidth{\linewidth}{%
\begin{tabular}{@{}l|cc|cc@{}}
\toprule
\multirow{2}{*}{Location} & \multicolumn{2}{c|}{L2 (m)$\downarrow$} & \multicolumn{2}{c}{Col. (\%)$\downarrow$} \\
& Before & TTA & Before & TTA \\
\midrule
Boston (Clean) & 0.61 & \textemdash & 0.01 & \textemdash \\
\midrule
Holland Village & 1.32 & \textbf{1.25}{\scriptsize (\textcolor{red}{+9.9\%})} & 0.51 & \textbf{0.47}{\scriptsize (\textcolor{red}{+8.0\%})} \\
One-North & 1.06 & \textbf{0.90}{\scriptsize (\textcolor{red}{+35.6\%})} & 0.45 & \textbf{0.25}{\scriptsize (\textcolor{red}{+45.5\%})} \\
Queenstown & 1.08 & \textbf{0.98}{\scriptsize (\textcolor{red}{+21.3\%})} & 0.10 & \textbf{0.06}{\scriptsize (\textcolor{red}{+44.4\%})} \\
\bottomrule
\end{tabular}}
\end{table}

\subsection{Strong-Interaction Scenarios}
Following \citet{jiang2025naturalistic}, we mine strong-interaction scenes comprising less than $5\%$ of Singapore test data. The Boston-trained SparseDrive model is evaluated on clean inputs and four visual corruptions.

Table~\ref{app:tab:ood-interaction} shows gains under every input condition. Averaging scores over the four corruptions, L2 and collision rate decrease by $7.3\%$ and $49.8\%$ relative to the matched unadapted scores.

\begin{table}[!htbp]
\setlength{\tabcolsep}{1.5pt}
\caption{\textcolor{black}{\textbf{Planning on strong-interaction scenes.} Before and TTA use the same input condition; AVG excludes Clean.}}
\label{app:tab:ood-interaction}
\centering\small
\setlength{\tabcolsep}{3pt}
\fitwidth{\linewidth}{%
\begin{tabular}{@{}l|cc|cc@{}}
\toprule
\multirow{2}{*}{Condition} & \multicolumn{2}{c|}{L2 (m)$\downarrow$} & \multicolumn{2}{c}{Col. (\%)$\downarrow$} \\
& Before & TTA & Before & TTA \\
\midrule
Clean & 0.84 & \cellcolor{blue!5}\textbf{0.78} & 1.15 & \cellcolor{blue!5}\textbf{0.29} \\
Gaussian & 0.98 & \cellcolor{blue!5}\textbf{0.95} & 1.09 & \cellcolor{blue!5}\textbf{0.44} \\
Fog & 0.86 & \cellcolor{blue!5}\textbf{0.74} & 0.51 & \cellcolor{blue!5}\textbf{0.09} \\
Snow & 0.93 & \cellcolor{blue!5}\textbf{0.86} & 1.40 & \cellcolor{blue!5}\textbf{0.89} \\
Brightness & 0.81 & \cellcolor{blue!5}\textbf{0.77} & 0.19 & \cellcolor{blue!5}\textbf{0.18} \\
\midrule
AVG & \textbf{0.90} & \cellcolor{blue!5}\textbf{0.83} & \textbf{0.80} & \cellcolor{blue!5}\textbf{0.40} \\
\bottomrule
\end{tabular}}
\end{table}

\section{Sensitivity to the Loss Weight}\label{app:d}

We test $\lambda\in\{0.1,0.3,1,3,10\}$ in Eq.~\ref{eq:joint_objective} on UniAD under severity-5 Gaussian noise, fixing other settings. All tested weights improve both metrics; $\lambda=1$ gives the lowest L2 and tied lowest collision rate, while $\lambda=10$ weakens the gains (Table~\ref{app:tab:loss-weight}).

\begin{table}[!htbp]
\setlength{\tabcolsep}{1.5pt}
\caption{\textcolor{black}{\textbf{Sensitivity to the perception--planning loss weight on UniAD.} Bold marks the best reported scores.}}
\label{app:tab:loss-weight}
\centering\small
\setlength{\tabcolsep}{6pt}
\begin{tabular}{@{}l|cc@{}}
\toprule
Setting & L2 (m)$\downarrow$ & Col. (\%)$\downarrow$ \\
\midrule
Without TTA & 1.38 & 0.95 \\
\midrule
$\lambda=0.1$ & 1.28 & \textbf{0.72} \\
$\lambda=0.3$ & 1.27 & \textbf{0.72} \\
$\lambda=1$ & \textbf{1.25} & \textbf{0.72} \\
$\lambda=3$ & 1.27 & 0.82 \\
$\lambda=10$ & 1.32 & 0.93 \\
\bottomrule
\end{tabular}
\end{table}

Across five architectures, $\lambda=1$ is best in seven of ten model--metric combinations, and $\lambda=3$ in the remaining three (Table~\ref{app:tab:loss-weight-models}), supporting $1$ as a shared default.

\begin{table}[!htbp]
\setlength{\tabcolsep}{1.5pt}
\caption{\textcolor{black}{\textbf{Best loss weight across architectures.} Each column gives the best tested weight for its metric.}}
\label{app:tab:loss-weight-models}
\centering\small
\setlength{\tabcolsep}{4pt}
\fitwidth{\linewidth}{%
\begin{tabular}{@{}l|cc@{}}
\toprule
Model & Best $\lambda$ (L2) & Best $\lambda$ (Col.) \\
\midrule
ST-P3~\citep{hu2022st} & 1 & 1 \\
UniAD~\citep{hu2023planning} & 1 & 1 \\
DistillDrive~\citep{yu2025distilldrive} & 3 & 3 \\
SparseDrive~\citep{sun2025sparsedrive} & 3 & 1 \\
DiffusionDrive~\citep{liao2025diffusiondrive} & 1 & 1 \\
\bottomrule
\end{tabular}}
\end{table}

\section{\textcolor{black}{Extended Component Ablation Studies}}\label{app:e}
We extend the main UniAD ablation (Table~\ref{tab:ablation}) to ST-P3 and SparseDrive under severity-5 Gaussian noise.

\subsection{Ablation on ST-P3}
On ST-P3, the joint objective improves both metrics over either loss alone. Stabilization further raises L2 recovery from $27.9\%$ to $34.3\%$, with $31.3\%$ collision recovery (Table~\ref{app:tab:ablation-stp}).

\begin{table}[!htbp]
\setlength{\tabcolsep}{1.5pt}
\caption{{\textbf{Component ablation on ST-P3.} Notation follows Table~\ref{tab:ablation}.}}
\centering
\fitwidth{\linewidth}{
\small
{\color{black}
\begin{tabular}{ccc|cccc|cccc}
\toprule
\multicolumn{3}{c|}{\textbf{Component}} & \multicolumn{4}{c|}{\textbf{L2 (m)↓}} & \multicolumn{4}{c}{\textbf{Col. (\%)↓}}  \\

$\mathcal{L}_{mm}$ & $\mathcal{L}_{temp}$ & $\mathcal{S}_t$ & 1s & 2s & 3s & \textbf{AVG} & 1s & 2s & 3s & \textbf{AVG} \\
\midrule
- & - & - & 2.51 & 5.22 & 8.13 & \textbf{5.29} & 0.89 & 5.08 & 9.42 & \textbf{5.13}  \\
\checkmark & - & - & 2.39 & 4.74 & 7.16 & \textbf{4.76}{\scriptsize (\textcolor{red}{+26.4\%})} & 0.66 & 4.13 & 7.66 & \textbf{4.15}{\scriptsize (\textcolor{red}{+27.8\%})}  \\

- & \checkmark & - & 2.39 & 4.74 & 7.19 & \textbf{4.77}{\scriptsize (\textcolor{red}{+25.9\%})} & 0.71 & 3.96 & 7.93 & \textbf{4.20}{\scriptsize (\textcolor{red}{+26.4\%})}  \\

\checkmark & \checkmark & - & 2.38 & 4.70 & 7.12 & \textbf{4.73}{\scriptsize (\textcolor{red}{+27.9\%})} & 0.66 & 3.86 & 7.49 & \textbf{4.00}{\scriptsize (\textcolor{red}{+32.1\%})}  \\

\cellcolor{blue!5}\checkmark & \cellcolor{blue!5}\checkmark & \cellcolor{blue!5}\checkmark & \cellcolor{blue!5}2.29 & \cellcolor{blue!5}4.57 & \cellcolor{blue!5}6.95 & \cellcolor{blue!5}\textbf{4.60}{\scriptsize (\textcolor{red}{+34.3\%})} & \cellcolor{blue!5}0.46 & \cellcolor{blue!5}3.88 & \cellcolor{blue!5}7.76 & \cellcolor{blue!5}\textbf{4.03}{\scriptsize (\textcolor{red}{+31.3\%})}\\
\bottomrule
\end{tabular}
}
}
\label{app:tab:ablation-stp}
\end{table}

\subsection{Ablation on SparseDrive}
Table~\ref{app:tab:ablation-sparse} examines objective combinations on SparseDrive. The full method improves both average metrics over the joint-loss variant, achieving $26.7\%$ L2 recovery and $27.3\%$ collision recovery.

\begin{table}[!htbp]
\caption{\textbf{Component ablation on SparseDrive under severity-5 Gaussian noise.} Notation follows Table~\ref{tab:ablation}.}
\label{app:tab:ablation-sparse}
\centering\small
\setlength{\tabcolsep}{1.5pt}
\fitwidth{\linewidth}{%
\begin{tabular}{ccc|cccc|cccc}
\toprule
\multicolumn{3}{c|}{\textbf{Component}} & \multicolumn{4}{c|}{\textbf{L2 (m)↓}} & \multicolumn{4}{c}{\textbf{Col. (\%)↓}}  \\

$\mathcal{L}_{mm}$ & $\mathcal{L}_{temp}$ & $\mathcal{S}_t$  & 1s & 2s & 3s & \textbf{AVG} & 1s & 2s & 3s & \textbf{AVG} \\
\midrule
- & - & - & 0.36 & 0.73 & 1.25 & \textbf{0.78} & 0.06 & 0.27 & 0.90 & \textbf{0.41}  \\

\checkmark & - & - & 0.36 & 0.71 & 1.20 & \textbf{0.75}{\scriptsize (\textcolor{red}{+20.0\%})} & 0.06 & 0.22 & 0.77 & \textbf{0.35}{\scriptsize (\textcolor{red}{+18.2\%})}  \\

- & \checkmark & - & 0.36 & 0.71 & 1.19 & \textbf{0.75}{\scriptsize (\textcolor{red}{+20.0\%})} & 0.13 & 0.22 & 0.64 & \textbf{0.33}{\scriptsize (\textcolor{red}{+24.2\%})}  \\

\checkmark & \checkmark & - & 0.35 & 0.71 & 1.20 & \textbf{0.75}{\scriptsize (\textcolor{red}{+20.0\%})} & 0.06 & 0.23 & 0.76 & \textbf{0.35}{\scriptsize (\textcolor{red}{+18.2\%})}  \\

\cellcolor{blue!5}\checkmark & \cellcolor{blue!5}\checkmark & \cellcolor{blue!5}\checkmark & \cellcolor{blue!5}0.36 & \cellcolor{blue!5}0.70 & \cellcolor{blue!5}1.16 & \cellcolor{blue!5}\textbf{0.74}{\scriptsize (\textcolor{red}{+26.7\%})} & \cellcolor{blue!5}\textcolor{black}{0.10} & \cellcolor{blue!5}0.22 & \cellcolor{blue!5}\textcolor{black}{0.64} & \cellcolor{blue!5}\textcolor{black}{\textbf{0.32}}{\scriptsize (\textcolor{red}{+27.3\%})}  \\
\bottomrule
\end{tabular}
}
\end{table}

\subsection{Within-Scene Continual Adaptation}\label{app:long-horizon}
Table~\ref{app:tab:within-scene-l2} expands Figure~\ref{fig:long-horizon}, comparing the full method with removal of $\mathcal{S}_t$ while retaining both losses. Frames~1--5 have zero RR. 

\begin{table}[!htbp]
  \caption{\textbf{Teacher--student stabilization during within-scene continual adaptation.} L2 is in meters and RR is in percent. Bold marks better adapted scores; ties are unbolded. Dashes denote unavailable L2 values.}
  \label{app:tab:within-scene-l2}
  \centering\small
  \setlength{\tabcolsep}{3pt}
  \textbf{(a) UniAD}\par\smallskip
  \fitwidth{\linewidth}{%
  \begin{tabular}{@{}c|c|cc|cc@{}}
    \toprule
    \multirow{2}{*}{Frames} & \multirow{2}{*}{Noisy L2} & \multicolumn{2}{c|}{Full \methodname{}} & \multicolumn{2}{c}{w/o $\mathcal{S}_t$} \\
    & & L2$\downarrow$ & RR$\uparrow$ & L2$\downarrow$ & RR$\uparrow$ \\
    \midrule
    1--5 & 2.6554 & \cellcolor{blue!5}2.655 & \textcolor{red}{$0.00$} & 2.655 & \textcolor{red}{$0.00$} \\
    6--10 & 1.2891 & \cellcolor{blue!5}\textbf{1.228} & \textcolor{red}{\textbf{$16.57$}} & 1.245 & \textcolor{red}{$11.95$} \\
    11--15 & 1.3711 & \cellcolor{blue!5}\textbf{1.275} & \textcolor{red}{\textbf{$22.37$}} & 1.311 & \textcolor{red}{$14.10$} \\
    16--20 & 1.3905 & \cellcolor{blue!5}\textbf{1.280} & \textcolor{red}{\textbf{$21.93$}} & 1.324 & \textcolor{red}{$13.24$} \\
    21--25 & 1.3996 & \cellcolor{blue!5}\textbf{1.280} & \textcolor{red}{\textbf{$23.04$}} & 1.328 & \textcolor{red}{$13.69$} \\
    26--30 & 1.4308 & \cellcolor{blue!5}\textbf{1.286} & \textcolor{red}{\textbf{$26.97$}} & 1.338 & \textcolor{red}{$17.29$} \\
    31--35 & 1.4343 & \cellcolor{blue!5}\textbf{1.301} & \textcolor{red}{\textbf{$27.02$}} & 1.335 & \textcolor{red}{$20.07$} \\
    \bottomrule
  \end{tabular}}
  \par\medskip
  \textbf{(b) SparseDrive}\par\smallskip
  \fitwidth{\linewidth}{%
  \begin{tabular}{@{}c|c|cc|cc@{}}
    \toprule
    \multirow{2}{*}{Frames} & \multirow{2}{*}{Noisy L2} & \multicolumn{2}{c|}{Full \methodname{}} & \multicolumn{2}{c}{w/o $\mathcal{S}_t$} \\
    & & L2$\downarrow$ & RR$\uparrow$ & L2$\downarrow$ & RR$\uparrow$ \\
    \midrule
    1--5 & 1.7575 & \cellcolor{blue!5}1.757 & \textcolor{red}{$0.00$} & 1.757 & \textcolor{red}{$0.00$} \\
    6--10 & 1.5492 & \cellcolor{blue!5}\textbf{1.440} & \textcolor{red}{\textbf{$25.35$}} & 1.475 & \textcolor{red}{$17.30$} \\
    11--15 & 1.5653 & \cellcolor{blue!5}\textbf{1.361} & \textcolor{red}{\textbf{$45.14$}} & 1.406 & \textcolor{red}{$35.16$} \\
    16--20 & 1.4286 & \cellcolor{blue!5}\textbf{1.280} & \textcolor{red}{\textbf{$40.27$}} & 1.337 & \textcolor{red}{$24.75$} \\
    21--25 & 1.3325 & \cellcolor{blue!5}\textbf{1.283} & \textcolor{red}{\textbf{$25.83$}} & 1.385 & \textcolor{red}{$-27.53$} \\
    26--30 & 1.3984 & \cellcolor{blue!5}\textbf{1.276} & \textcolor{red}{\textbf{$42.17$}} & 1.377 & \textcolor{red}{$7.35$} \\
    31--35 & 1.4347 & \cellcolor{blue!5}\textbf{1.280} & \textcolor{red}{\textbf{$45.71$}} & 1.325 & \textcolor{red}{$32.54$} \\
    \bottomrule
  \end{tabular}}
  \par\medskip
  \textbf{(c) DistillDrive}\par\smallskip
  \fitwidth{\linewidth}{%
  \begin{tabular}{@{}c|c|cc|cc@{}}
    \toprule
    \multirow{2}{*}{Frames} & \multirow{2}{*}{Noisy L2} & \multicolumn{2}{c|}{Full \methodname{}} & \multicolumn{2}{c}{w/o $\mathcal{S}_t$} \\
    & & L2$\downarrow$ & RR$\uparrow$ & L2$\downarrow$ & RR$\uparrow$ \\
    \midrule
    1--5 & 0.900 & \cellcolor{blue!5}\textemdash & \textcolor{red}{$0.00$} & \textemdash & \textcolor{red}{$0.00$} \\
    6--10 & 0.783 & \cellcolor{blue!5}0.747 & \textcolor{red}{$24.22$} & \textbf{0.731} & \textcolor{red}{\textbf{$31.51$}} \\
    11--15 & 0.793 & \cellcolor{blue!5}\textbf{0.713} & \textcolor{red}{\textbf{$39.75$}} & 0.725 & \textcolor{red}{$33.88$} \\
    16--20 & 0.756 & \cellcolor{blue!5}\textbf{0.704} & \textcolor{red}{\textbf{$43.80$}} & 0.726 & \textcolor{red}{$33.84$} \\
    21--25 & 0.715 & \cellcolor{blue!5}\textbf{0.681} & \textcolor{red}{\textbf{$53.87$}} & 0.717 & \textcolor{red}{$37.59$} \\
    26--30 & 0.736 & \cellcolor{blue!5}\textbf{0.698} & \textcolor{red}{\textbf{$46.39$}} & 0.750 & \textcolor{red}{$22.84$} \\
    31--35 & 0.737 & \cellcolor{blue!5}\textbf{0.688} & \textcolor{red}{\textbf{$50.75$}} & 0.694 & \textcolor{red}{$47.98$} \\
    \bottomrule
  \end{tabular}}
\end{table}

\par\smallskip
Recovery fluctuates with driving conditions but generally increases during adaptation. The full method gives higher RR after frames~1--5 on UniAD and SparseDrive, and from frames~11--15 on DistillDrive. In SparseDrive frames~21--25, it achieves $25.83\%$ RR versus $-27.53\%$ without $\mathcal{S}_t$, illustrating reduced local degradation.

\section{\textcolor{black}{Implementation Details and Task Coupling}}\label{app:f}

\subsection{Scope of Normalization Adaptation}\label{app:ln-scope}
Only selected normalization scale and bias tensors are adapted: LayerNorm in transformer models and BatchNorm in ST-P3. Table~\ref{app:tab:normalization-budget} counts affine scalars, excluding running statistics, optimizer states, and EMA copies.

\begin{table}[!htbp]

  \caption{\textbf{Normalization parameter budgets for the profiled models.} Component counts are exact numbers of selected affine scalars. Shared/perception includes shared feature-processing modules. ST-P3 uses BatchNorm; the other rows use LayerNorm.}
  \label{app:tab:normalization-budget}
  \centering\footnotesize
  \setlength{\tabcolsep}{2.5pt}
  \fitwidth{\linewidth}{%
  \begin{tabular}{@{}l|rr|rr@{}}
    \toprule
    Model & Shared/perc. & Planning & Total & Table~\ref{tab:efficiency} \\
    \midrule
    UniAD & 25,088 & 5,120 & 30,208 & 30.21K \\
    ST-P3 & 256 & 928 & 1,184 & 1.18K \\
    DiffusionDrive & 6,144 & 2,560 & 8,704 & 8.70K \\
    SparseDrive & 6,144 & 1,536 & 7,680 & 7.68K \\
    DistillDrive & 6,144 & 1,536 & 7,680 & 7.68K \\
    CoIRL-AD & 0 & 1,536 & 1,536 & 1.54K \\
    \bottomrule
  \end{tabular}}
\end{table}

\subsection{Model-Specific Adaptation on nuScenes}\label{app:model-adapted-extensions}
All models use observed ego motion for temporal consistency; parameter counts are in Table~\ref{app:tab:normalization-budget}.

\textbf{UniAD} links perception and planning through BEV-query transformers. Entropy uses ten-class sigmoid scores from six classification branches; selected LayerNorm modules cover the perception transformer, ego-state embedding, classification branches, and planning attention/fusion.

\textbf{ST-P3} uses dense spatiotemporal BEV features. Entropy uses softmax vehicle/pedestrian segmentation probabilities; adaptation targets BatchNorm affine parameters in segmentation heads and planning channel-reduction blocks, with running statistics fixed.

\textbf{SparseDrive, DistillDrive, and nuScenes DiffusionDrive} use sparse perception with ego planning, multi-mode planning distillation, and diffusion trajectory decoding, respectively. Entropy uses sigmoid object scores from six detection refinement stages. Selected LayerNorm modules lie in detection classification branches, the planning-anchor encoder, and the respective planning classification branches.

\textbf{CoIRL-AD} combines imitation and reinforcement learning in a latent world model. Its imitation pathway lacks a trained probabilistic perception head, so adaptation uses temporal consistency alone. LayerNorm parameters in waypoint attention are adapted, with the reinforcement-learning branch frozen.

\subsection{Implementation Details of TTA Baselines}\label{app:baseline-details}
The main baselines share our objective and loss weight (Eq.~\ref{eq:joint_objective}), selected UniAD LayerNorm parameters, and per-scene reset rule. Perception entropy uses ten-class sigmoid scores from six classification branches, averaged over branches, queries, and classes; temporal consistency uses recomputed historical plans and observed ego motion.

Tent~\citep{wang2020tent} directly minimizes the shared objective. EATA~\citep{niu2022efficient} retains reliability/diversity filtering and EWC; EATA--E removes EWC. SAR~\citep{niu2023towards} retains reliability filtering, sharpness-aware updates, and model recovery. CoTTA~\citep{wang2022continual} retains its EMA teacher and stochastic source restoration.

\subsection{Baselines with Their Original Adaptation Objectives}\label{app:perception-only-baselines}
Table~\ref{app:tab:perception-only-baselines} evaluates the original perception-entropy objectives on UniAD under severity-5 Gaussian noise, without temporal planning consistency. Tent, EATA, EATA--E, and SAR adapt perception-branch LayerNorm parameters; EcoTTA adapts its lightweight perception modules. Each retains its own update mechanism.

The strongest recoveries are $9.1\%$ for L2 and $16.7\%$ for collisions, versus $15.8\%$ and $32.8\%$ under the shared perception--planning objective in Table~\ref{tab:comparison}. This comparison supports accounting for downstream planning during adaptation.

\begin{table}[!htbp]
\setlength{\tabcolsep}{1.5pt}
\caption{\textbf{TTA baselines with their original adaptation objectives on UniAD.} L2 error and collision rate at each prediction horizon under severity-5 Gaussian noise.}
\label{app:tab:perception-only-baselines}
\centering\small
\setlength{\tabcolsep}{3pt}
\fitwidth{\linewidth}{%
\begin{tabular}{@{}l|cccc|cccc@{}}
\toprule
\multirow{2}{*}{Method} & \multicolumn{4}{c|}{L2 (m)$\downarrow$} & \multicolumn{4}{c}{Col. (\%)$\downarrow$} \\
& 1s & 2s & 3s & AVG & 1s & 2s & 3s & AVG \\
\midrule
Without TTA & 0.60 & 1.29 & 2.25 & \textbf{1.38} & 0.81 & 0.83 & 1.23 & \textbf{0.95} \\
\midrule
Tent \citep{wang2020tent} & 0.58 & 1.28 & 2.24 & \textbf{1.37}{\scriptsize (\textcolor{red}{+2.3\%})} & 0.81 & 0.96 & 1.05 & \textbf{0.94}{\scriptsize (\textcolor{red}{+1.5\%})} \\
EcoTTA \citep{song2023ecotta} & 0.60 & 1.29 & 2.24 & \textbf{1.38}{\scriptsize (\textcolor{red}{0.0\%})} & 0.83 & 0.83 & 1.13 & \textbf{0.93}{\scriptsize (\textcolor{red}{+3.0\%})} \\
EATA \citep{niu2022efficient} & 0.55 & 1.26 & 2.23 & \textbf{1.35}{\scriptsize (\textcolor{red}{+6.8\%})} & 0.76 & 0.75 & 1.01 & \textbf{0.84}{\scriptsize (\textcolor{red}{+16.7\%})} \\
EATA--E \citep{niu2022efficient} & 0.53 & 1.25 & 2.23 & \textbf{1.34}{\scriptsize (\textcolor{red}{+9.1\%})} & 0.71 & 0.91 & 1.05 & \textbf{0.89}{\scriptsize (\textcolor{red}{+9.1\%})} \\
SAR \citep{niu2023towards} & 0.56 & 1.27 & 2.24 & \textbf{1.36}{\scriptsize (\textcolor{red}{+4.5\%})} & 0.76 & 0.78 & 1.05 & \textbf{0.86}{\scriptsize (\textcolor{red}{+13.6\%})} \\
\bottomrule
\end{tabular}}
\end{table}

\section{Case Study and Error Correlation Analysis}\label{app:g}

\subsection{Qualitative Success and Failure Cases}

\begin{figure}[!htbp]
  \centering
  \includegraphics[width=\linewidth,height=.52\textheight,keepaspectratio]{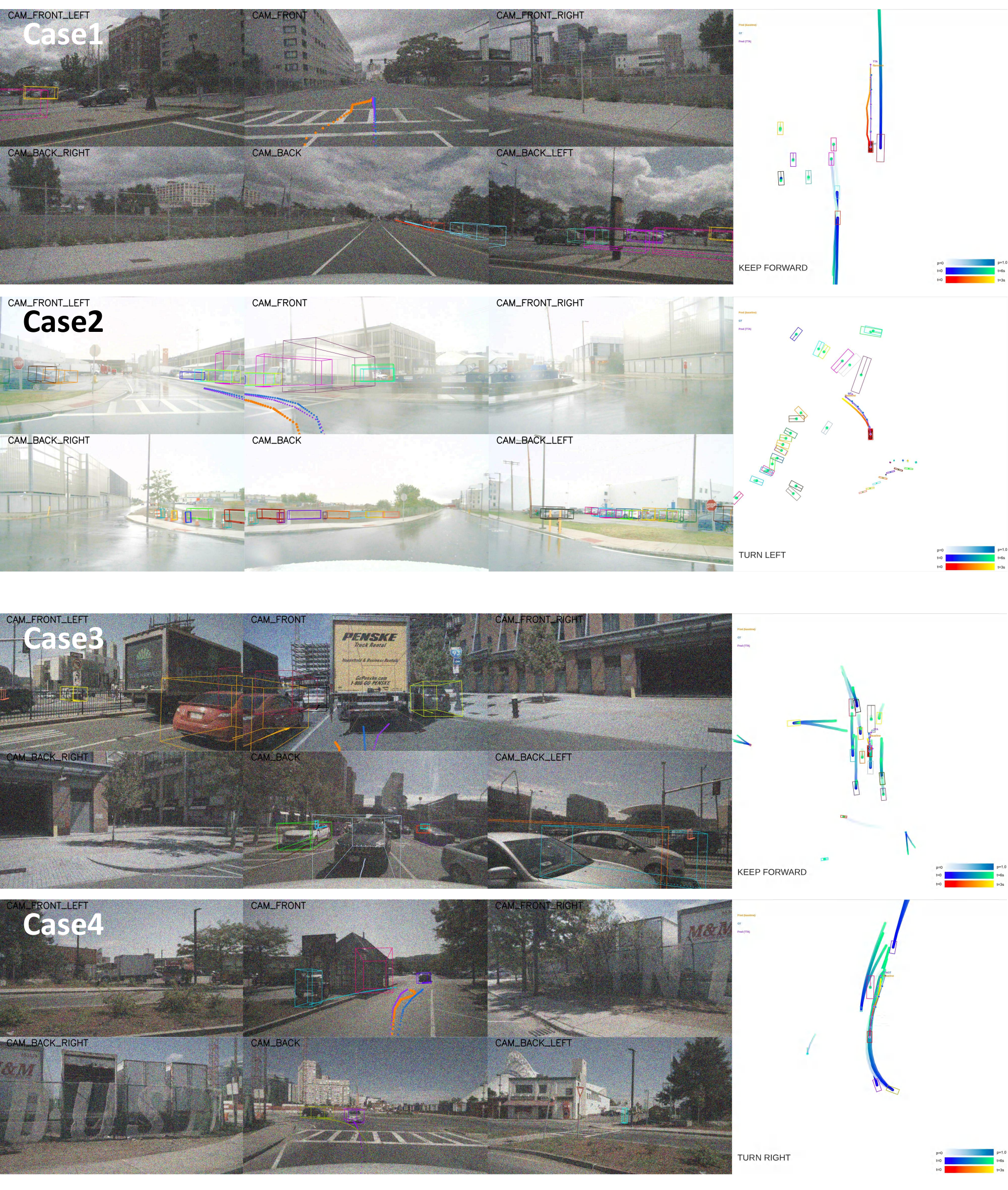}
  \caption{\textbf{Qualitative trajectory comparisons before and after TTA.} Ground Truth (blue), Before TTA (orange), and After TTA (purple) denote the reference, pretrained, and adapted trajectories, respectively. Cases 1--2 illustrate successful corrections; Cases 3--4 show remaining failures.}
  \label{fig:app-case-vis}
\end{figure}

In Figure~\ref{fig:app-case-vis}, Cases~1--2 show closer alignment with ground truth under Gaussian noise and brightness corruption. Case~3, using a non-default configuration, overshoots to the opposite side of the reference; Case~4 becomes smoother but has higher L2 error, showing that smoothness and displacement accuracy can disagree.

\subsection{Scenario-Level Error Analysis}
\begin{figure}[t]
  \centering
  \includegraphics[width=.96\linewidth,trim=0bp 0bp 480.025bp 0bp,clip]{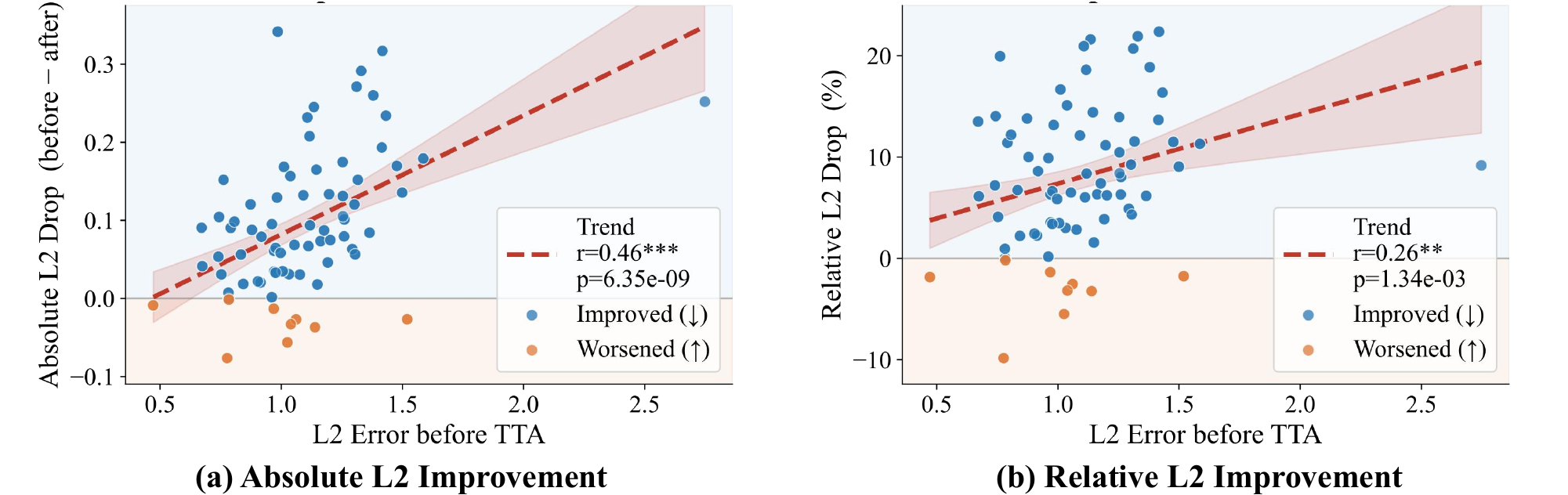}
  \par\smallskip
  \includegraphics[width=.96\linewidth,trim=480.025bp 0bp 0bp 0bp,clip]{pdf/app-case.pdf}
  \caption{\textbf{Scenario-level L2 changes under severe Gaussian noise.} Absolute (a) and relative (b) reductions are plotted against the unadapted L2 error. Blue and orange points denote improved and degraded scenarios, respectively; fitted trends and correlation statistics summarize the relationship with initial error.}
  \label{fig:app-case}
\end{figure}

Figure~\ref{fig:app-case} relates absolute and relative L2 reductions to unadapted errors under severity-5 Gaussian noise. Larger initial errors tend to accompany larger reductions, although some scenarios deteriorate after adaptation.

\section{Closed-Loop Evaluation on Bench2Drive}\label{app:b2d-results}
Table~\ref{app:tab:b2d-results} evaluates five models on the ten-route Bench2Drive~\citep{jia2024bench2drive} DEV10 subset under severity-5 Gaussian noise. Clean and Noisy are unadapted controls on matched routes; TTA uses corrupted inputs. When several configurations are available, we report the one with highest adapted Driving Score.

Driving Score (DS) combines route completion and infraction penalties; SR and Col. are percentages of successful routes and routes with collisions. RR follows Appendix~\ref{app:main-details}; dashes indicate zero clean-to-noisy gaps.

\begin{figure*}[!t]
  \centering
  \includegraphics[page=1,width=\textwidth]{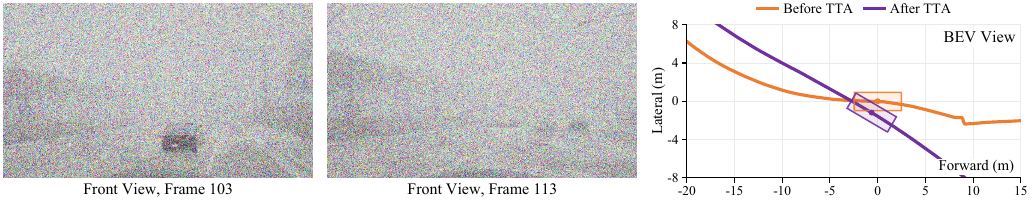}
  \par\smallskip
  \includegraphics[page=2,width=\textwidth]{pdf/b2d_additional_closed_loop_cases.pdf}
  \caption{\textbf{Additional closed-loop cases on Bench2Drive.} Top: pedestrian crossing. Bottom: road departure. Each row shows two front-camera frames and the executed trajectories in BEV. Orange and purple denote Before TTA and After TTA, respectively.}
  \label{fig:app-b2d-additional-cases}
\end{figure*}

\begin{table}[t]
\caption{\textbf{Closed-loop results on Bench2Drive.} Results on the ten-route DEV10 subset under severity-5 Gaussian noise.}
\label{app:tab:b2d-results}
\centering\small
\setlength{\tabcolsep}{4pt}
\renewcommand{\arraystretch}{1.02}
\fitwidth{\linewidth}{%
\begin{tabular}{@{}l|ccc@{}}
\toprule
Setting & DS$\uparrow$ & SR (\%)$\uparrow$ & Col. (\%)$\downarrow$ \\
\midrule
\multicolumn{4}{@{}l}{\textbf{DriveTransformer}} \\
Clean & 55.34 & 30 & 50 \\
Noisy & 38.74 & 30 & 70 \\
\rowcolor{blue!5}TTA & \textbf{43.74}{\scriptsize (\textcolor{red}{+30.09\%})} & \textbf{30}{\scriptsize (---)} & \textbf{70}{\scriptsize (\textcolor{red}{0\%})} \\
\specialrule{0.3pt}{0.3pt}{0.3pt}
\multicolumn{4}{@{}l}{\textbf{TCP-traj}} \\
Clean & 54.56 & 20 & 60 \\
Noisy & 52.14 & 20 & 70 \\
\rowcolor{blue!5}TTA & \textbf{56.01}{\scriptsize (\textcolor{red}{+159.88\%})} & \textbf{30}{\scriptsize (---)} & \textbf{60}{\scriptsize (\textcolor{red}{+100\%})} \\
\specialrule{0.3pt}{0.3pt}{0.3pt}
\multicolumn{4}{@{}l}{\textbf{SparseDriveV2}} \\
Clean & 82.57 & 60 & 20 \\
Noisy & 40.56 & 10 & 80 \\
\rowcolor{blue!5}TTA & \textbf{45.28}{\scriptsize (\textcolor{red}{+11.25\%})} & \textbf{10}{\scriptsize (\textcolor{red}{0\%})} & \textbf{80}{\scriptsize (\textcolor{red}{0\%})} \\
\specialrule{0.3pt}{0.3pt}{0.3pt}
\multicolumn{4}{@{}l}{\textbf{TransFuser++}} \\
Clean & 73.42 & 40 & 30 \\
Noisy & 50.01 & 20 & 70 \\
\rowcolor{blue!5}TTA & \textbf{59.34}{\scriptsize (\textcolor{red}{+39.85\%})} & \textbf{20}{\scriptsize (\textcolor{red}{0\%})} & \textbf{40}{\scriptsize (\textcolor{red}{+75.00\%})} \\
\specialrule{0.3pt}{0.3pt}{0.3pt}
\multicolumn{4}{@{}l}{\textbf{SimLingo}} \\
Clean & 89.00 & 70 & 20 \\
Noisy & 68.63 & 40 & 40 \\
\rowcolor{blue!5}TTA & \textbf{74.24}{\scriptsize (\textcolor{red}{+27.56\%})} & \textbf{60}{\scriptsize (\textcolor{red}{+66.67\%})} & \textbf{30}{\scriptsize (\textcolor{red}{+50.00\%})} \\
\bottomrule
\end{tabular}}
\end{table}

DS improves on all five models. TransFuser++ reduces collision rate from $70\%$ to $40\%$, and SimLingo improves DS, SR, and collision rate. These results complement the UniAD cut-in case in Figure~\ref{fig:b2d-cutin-case} with quantitative closed-loop evidence.

Figure~\ref{fig:app-b2d-additional-cases} presents two additional closed-loop cases on Bench2Drive. In the pedestrian-crossing case, \methodname{} enables earlier deceleration to yield to the crossing pedestrian. In the road-departure case, the unadapted vehicle leaves the road, whereas the adapted vehicle remains within the road boundaries. These examples illustrate corrections in both interaction handling and road adherence during closed-loop execution.

\FloatBarrier
\section{\textcolor{black}{Limitations and Future Work}}\label{app:h}
Future work will study when to trigger, pause, and resume adaptation using unlabeled signals, and how to adjust update step sizes dynamically to balance adaptation speed and stability under changing conditions.

\end{document}